%% file: main.tex
\documentclass[nohyperref]{article}

\usepackage[utf8]{inputenc} 
\usepackage[T1]{fontenc}    
\PassOptionsToPackage{hyphens}{url}
\usepackage{url}            
\usepackage{booktabs}       
\usepackage{nicefrac}       
\usepackage{microtype}      
\usepackage{appendix}
\usepackage[skip=5pt,font=small]{caption}  
\usepackage{subcaption}  
\usepackage{paralist}
\usepackage{xurl} 

\usepackage{wrapfig}
\usepackage{colortbl}
\usepackage{makecell}
\usepackage{amsfonts}       
\usepackage{amsmath}
\usepackage{amssymb}
\usepackage{mathtools}
\usepackage{amsthm}
\usepackage{hyperref}

\usepackage{todonotes}
\usepackage{multirow}
\usepackage[linesnumbered,ruled, vlined]{algorithm2e}

\usepackage[accepted]{arxiv2022}

\usepackage[capitalize,noabbrev]{cleveref}

\theoremstyle{plain}

\theoremstyle{definition}

\theoremstyle{remark}

\input{packages}

\input{commands}

\input{glossary}

\Crefformat{figure}{#2Fig.~#1#3}
\Crefmultiformat{figure}{Figs.~#2#1#3}{ and~#2#1#3}{, #2#1#3}{ and~#2#1#3}
\Crefformat{table}{#2Tab.~#1#3}
\Crefmultiformat{table}{Tabs.~#2#1#3}{ and~#2#1#3}{, #2#1#3}{ and~#2#1#3}

\crefformat{equation}{(#2#1#3)}

\makeatletter
\renewcommand\paragraph{\@startsection{paragraph}{4}{\z@}%
{0.5ex \@plus.1ex \@minus.1ex}%
{-1em}%
{\normalfont\normalsize\bfseries}}
\makeatother

\arxivtitlerunning{Adversarial Masking for Self-Supervised Learning}

\begin{document}

\twocolumn[
\arxivtitle{Adversarial Masking for Self-Supervised Learning}




\begin{arxivauthorlist}
\arxivauthor{Yuge Shi}{oxford}
\arxivauthor{N.\ Siddharth}{sid}
\arxivauthor{Philip H.S.\ Torr}{oxford}
\arxivauthor{Adam R.\ Kosiorek}{deepmind}

\end{arxivauthorlist}

\arxivaffiliation{oxford}{University of Oxford}
\arxivaffiliation{sid}{The University of Edinburgh \& The Alan Turing Institute}
\arxivaffiliation{deepmind}{DeepMind}

\arxivcorrespondingauthor{Yuge Shi}{yshi@robots.ox.ac.uk}
\arxivcorrespondingauthor{Adam Kosiorek}{adamrk@deepmind.com}

\arxivkeywords{Machine Learning, self-supervised learning, representation learning}

\vskip 0.3in
]



\printAffiliationsAndNotice{}  

\input{0abstract}

\input{1introductionv4}

\input{2methodsv4}
\input{3experiments}
\input{4relatedwork}

\input{5conclusion}
\input{6acknowledgement}

\newpage
\bibliography{references}
\bibliographystyle{arxiv2022}

\newpage
\input{6appendix1}

\end{document}

%% file: packages.tex
\DontPrintSemicolon

\definecolor{function}{HTML}{BEA31E}
\definecolor{for}{HTML}{EE6C4D}
\definecolor{comment}{HTML}{80854D}
\definecolor{lightgreen}{HTML}{F2ECE3}
\definecolor{lightblue}{HTML}{D1EDFF}
\definecolor{darkgreen}{HTML}{72A276}
\definecolor{grey}{HTML}{666666}
\SetAlFnt{\ttfamily}

\SetKwFor{For}{}{}{}%

\SetKwComment{Comment}{\color{comment}$\#$ }{}
\SetKw{KwIn}{\color{function}in}
\SetKw{KwFor}{\color{function}for}
\SetKw{Kwadios}{\color{for}adios}
\SetKw{Kwsimclr}{\color{for}simclr}

\SetKwProg{Function}{def}{}{}

\SetKw{Continue}{continue}

\usepackage{bm} 
\usepackage{enumitem} 

%% file: commands.tex
\newcommand{\x}{\bm{x}}
\newcommand{\m}{\bm{m}}
\newcommand{\z}{\bm{z}}
\newcommand{\h}{\bm{h}}
\newcommand{\y}{\bm{y}}

\newcommand{\M}{\mathcal{M}}
\newcommand{\I}{\mathcal{I}}
\newcommand{\RR}{\mathbb{R}}
\newcommand{\loss}{\mathcal{L}}

\newcommand{\std}[1]{(\textit{\scriptsize$\pm$\scriptsize #1})}
\newcommand{\sci}[2]{#1e-#2}

\definecolor{pink}{HTML}{ff00ff}
\definecolor{orange}{HTML}{ff9900}
\definecolor{blue}{HTML}{0000ff}

\newcommand*\circled[1]{{\scriptsize \tikz[baseline=(char.base)]{%
            \node[shape=circle,draw,inner sep=1pt] (char) {#1};}}}

%% file: glossary.tex
\usepackage[acronym,smallcaps,nowarn,section,nogroupskip,nonumberlist]{glossaries}

\glsdisablehyper 

\newacronym{VAE}{VAE}{variational autoencoder}
\newacronym{VAEAC}{vaeac}{variational autoencoder with arbitrary conditioning}
\newacronym{SSL}{SSL}{self-supervised learning}
\newacronym{IS}{is}{importance sampling}
\newacronym{IWAE}{iwae}{importance-weighted auto-encoder}
\newacronym{FIVO}{fivo}{Filtering Variational Objectives}
\newacronym{SMC}{smc}{sequential Monte Carlo}
\newacronym{SSM}{ssm}{state-space model}
\newacronym{SGA}{sga}{stochastic gradient ascent}
\newacronym{SGD}{sgd}{stochastic gradient descent}
\newacronym{ELBO}{elbo}{evidence lower bound}
\newacronym{KL}{kl}{Kullback-Leibler}
\newacronym{LSTM}{lstm}{long short-term memory}
\newacronym{CNN}{cnn}{convolutional neural network}
\newacronym{MML}{mml}{maximum marginal likelihood}

\newacronym{REINFORCE}{reinforce}{REINFORCE}
\glsunset{REINFORCE}
\newacronym{RL}{rl}{reinforcement learning}

\newacronym{ADAM}{adam}{ADAM}
\glsunset{ADAM}
\newacronym{RMSprop}{rmsprop}{RMSprop}
\glsunset{RMSprop}

\newacronym{GAN}{gan}{generative adversarial network}
\newacronym{GRU}{gru}{gated recurrent unit}
\newacronym{MLP}{mlp}{multilayer perceptron}
\newacronym{MC}{mc}{Monte Carlo}
\newacronym{mibp}{mIBP}{Markov Indian Buffer Process}
\newacronym{VRNN}{vrnn}{Variational Recurrent Neural Network}
\newacronym{LGSSM}{lgssm}{linear Gaussian state space model}
\newacronym[firstplural=recurrent neural networks, plural=RNNs]{RNN}{rnn}{recurrent neural network}

\newacronym{ELU}{elu}{exponential linear unit}

\newacronym{MNIST}{mnist}{mnist}
\glsunset{MNIST}

\newacronym{BERT}{BERT}{bidirectional encoder representations from transformers}
\newacronym{NLP}{NLP}{natural language processing}
\newacronym{MAE}{MAE}{masked autoencoder}
\newacronym{BEiT}{BEiT}{Bidirectional Encoder representation from Image Transformers}
\newacronym{iBOT}{iBOT}{Image BERT pre-training with Online Tokeniser}
\newacronym{ADIOS}{ADIOS}{adversarial inpainter occluder self-supervision}
\newacronym{ViT}{ViT}{Vision Transformer}
\newacronym{MLM}{MLM}{masked language models}
\newacronym{MIM}{MIM}{masked image modeling}
\newacronym{SimCLR}{SimCLR}{-}
\newacronym{SimSiam}{SimSiam}{simple siamese network}
\newacronym{BYOL}{BYOL}{bootstrap your own latent}

\newcommand{\bert}{BERT}

\newcommand{\mae}{{MAE}}
\newcommand{\beit}{{BEiT}}

\newcommand{\adios}{{ADIOS}}
\newcommand{\vit}{{ViT}}

\newcommand{\mim}{{MIM}}

\newcommand{\knn}{$k$-NN}

\newacronym{MONET}{monet}{monet}
\glsunset{MONET}
\newacronym{GENESIS}{genesis}{genesis}
\glsunset{GENESIS}

\newacronym{GECO}{geco}{generalized ELBO with constrained optimization}

%% file: 0abstract.tex
\begin{abstract}
    We propose ADIOS, a \gls{MIM} framework for self-supervised learning, which simultaneously learns a masking function and an image encoder using an adversarial objective.
    The image encoder is trained to minimise the distance between representations of the original and that of a masked image. The masking function, conversely, aims at maximising this distance.
    ADIOS consistently improves on state-of-the-art self-supervised learning (SSL) methods on a variety of tasks and datasets---including classification on ImageNet100 and STL10, transfer learning on CIFAR10/100, Flowers102 and iNaturalist, as well as robustness evaluated on the backgrounds challenge \citep{bgchallenge}---while generating semantically meaningful masks.
    Unlike modern MIM models such as MAE, BEiT and iBOT, ADIOS does not rely on the image-patch tokenisation construction of Vision Transformers, and can be implemented with convolutional backbones.
    We further demonstrate that the masks learned by ADIOS are more effective in improving representation learning of SSL methods than masking schemes used in popular MIM models.
    Code is available at \href{https://github.com/YugeTen/adios}{\texttt{https://github.com/YugeTen/adios}}.
\end{abstract}


%% file: 1introductionv4.tex
\section{Introduction} \label{sec:intro}

\begin{figure}[t]
  \includegraphics[width=\linewidth]{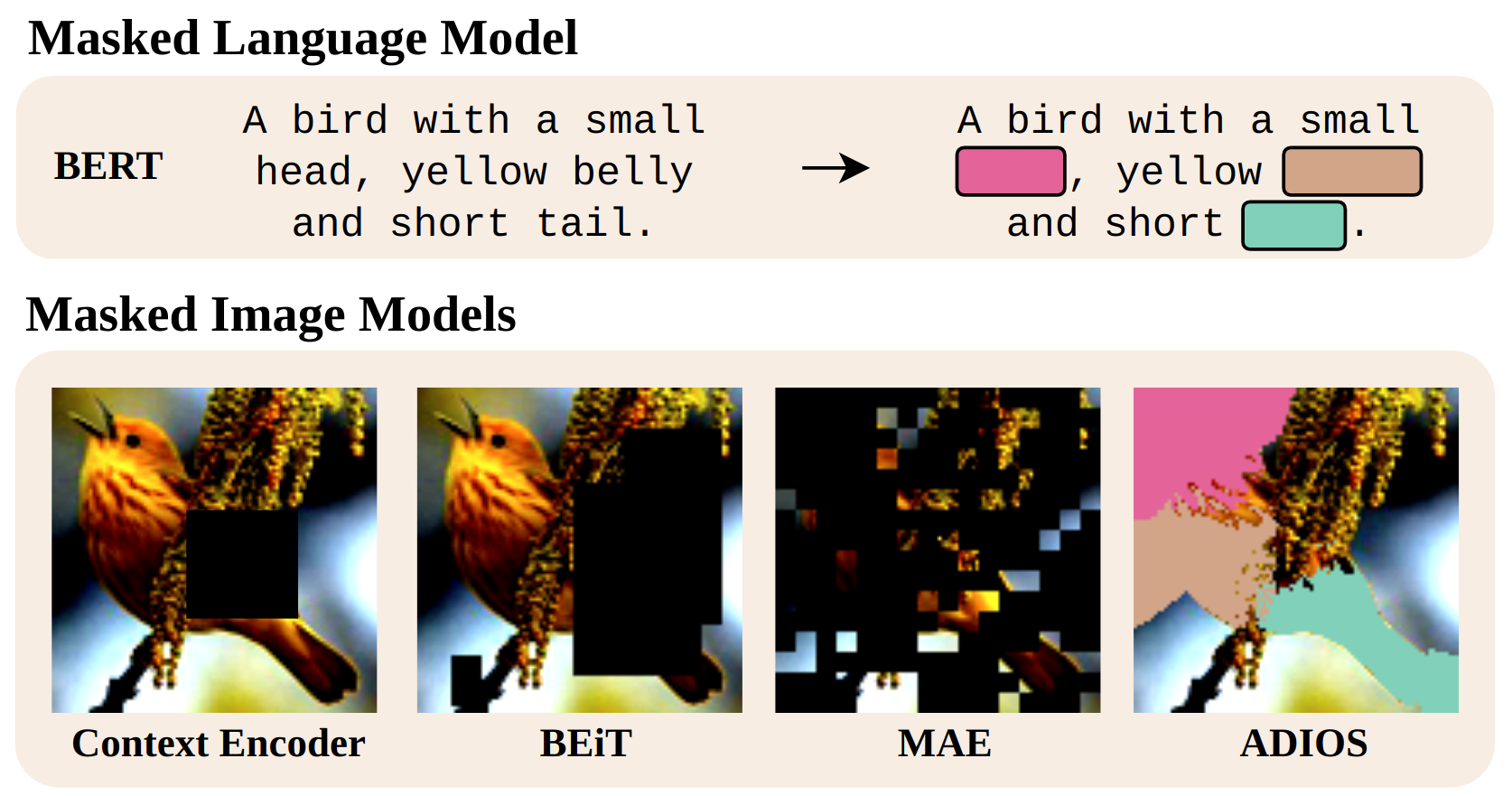}
  \vspace*{-1.2\baselineskip}
  \caption{
    Self-supervised language, and vision, models learn representations by imputing data removed by masking.
    \emph{\textbf{BERT}}: random word masks; \emph{\textbf{Context encoder}}: random, fix-shaped mask; \emph{\textbf{BEiT}}: random `blockwise' masking; \emph{\textbf{MAE}}: randomly mask out $75\%$ of the image; \emph{\textbf{ADIOS}}: multiple masks (N=3) generated by an adversarially trained masking model, post-processed with fully connected conditional random fields \citep{crf}.
  }
  \label{fig:hook}
  \vspace*{-1.2\baselineskip}
\end{figure}

The goal of \Acrfull{MIM} is to learn image representations, in a self-supervised fashion, by occluding parts of the input images.
\Gls{MIM} is inspired by significant advances in natural language modelling such as \bert\ \citep{devlin2018bert}, where the model is trained to fill-in words randomly removed from a sentence (\Cref{fig:hook}, top row).
Recent work, including \mae\ \citep{he2021mae} and \beit\ \citep{bao2021beit}, show that these gains are at least partially transferable to vision.
The task of a \gls{MIM} model is therefore similar to \bert, e.g., given an image of a bird in \Cref{fig:hook}, it needs to reason about what the bird might be sitting on or what colour the bird's belly is given the visible context (bottom row).
However, while missing words describe \emph{whole} semantic entities (e.g. ``head''), the masks used for context encoder (\citet{Pathak2016context}, which pioneered \gls{MIM}), \beit\ and \mae\, typically have no such constraint (\Cref{fig:hook} bottom, left to right).
Imputation under such schemes is conceptually simpler, as random masking only \emph{partially} obscures meaningful visual entities, which allows easier inference of missing values by leveraging strong correlations at the local-pixel level\footnote{Akin to randomly masking letters in a sentence for \acrshort{NLP}.}.

To narrow the gap between pixel masking and word masking, we posit that one needs to occlude \emph{whole entities in the image}. 
This encourages the model to perform imputation by complex semantic reasoning using the unmasked context (e.g. given a bird with a yellow body, it is likely to have a yellow head) rather than leveraging simple local correlations, which can benefit representation learning.
%
Interestingly, \citet{he2021mae} are motivated by similar hypothesis and propose to occlude a large fraction (up to $75\%$) of the image, removing complete entities by a higher chance, which they find is essential for good performance.
Here, we suggest that it is actually \emph{what is masked}, not so much \emph{how much is masked}, that is crucial for effective self-supervised representation learning.

To this end, we investigate {learning to mask} with an adversarial objective, where an {occlusion model} is asked to make reasoning about missing parts of the scene more difficult.
This novel representation-learning algorithm, called \textbf{Ad}versarial \textbf{I}nference-\textbf{O}cclusion \textbf{S}elf-supervision (\adios), can identify and mask out regions of correlated pixels within an image (\Cref{fig:hook}, bottom right), which brings it closer to the word-masking regime in natural language.
And as we shall see in \Cref{sec:emp_eval}, it consistently improves performance of state-of-the-art \gls{SSL} algorithms.

Some \gls{MIM} methods employ a generative component for representation learning, by learning to reconstruct the masked image.
However, it has been shown \cite{bao2021beit,ramesh2021zero} that pixel-level reconstruction tasks waste modelling capacity on high-frequency details over low-frequency structure, leading to subpar performance.
We hence frame \adios\ as an encoder-only framework that minimises the distance between the \emph{representations} of the original image and the masked image.
The occlusion model, which is trained adversarially to the encoder, tries to minimises this same distance.
We further discuss in \Cref{sec:ae_imputer} that, compared to the generative setup, the encoder-only setup optimises a functionally superior objective for representation learning.
Note that the encoder objective is compatible with many recent augmentation-based Siamese \acrlong{SSL} (\acrshort{SSL}; \citet{chen2020simclr, chen2021simsiam}) methods.
We show that \adios\ consistently improves performance of these \acrshort{SSL} objectives, showcasing the generality of our approach.

Our main contributions are as follows,
\begin{enumerate}[topsep=-0.5em,itemsep=0em,partopsep=0em,parsep=0em,leftmargin=*]
    \item A novel adversarial Siamese-style \gls{MIM} framework, that unlike other \mim\ methods is not limited to using \vit\ as backbone---advantageous given recent discoveries of modernised-convnet superiority over \vit s \citep{2022convnet,touvron2021augmenting}; 
    \item Qualitative and quantitative analyses showing that masks generated by \adios\ are semantically meaningful;
    \item Analysis of how different masking schemes affect representation-learning performance of \gls{SSL} models.
    We find models trained with ADIOS and ground-truth object masks significantly outperform other masking schemes/no mask, demonstrating the efficacy of semantically meaningful masks for representation learning. 
\end{enumerate}

%% file: 2methodsv4.tex
\section{Methodology}\label{sec:methods}

\begin{wrapfigure}[9]{r}{0.55\linewidth}
  \centering
  \vspace*{-1.5\baselineskip}
  \includegraphics[width=0.95\linewidth]{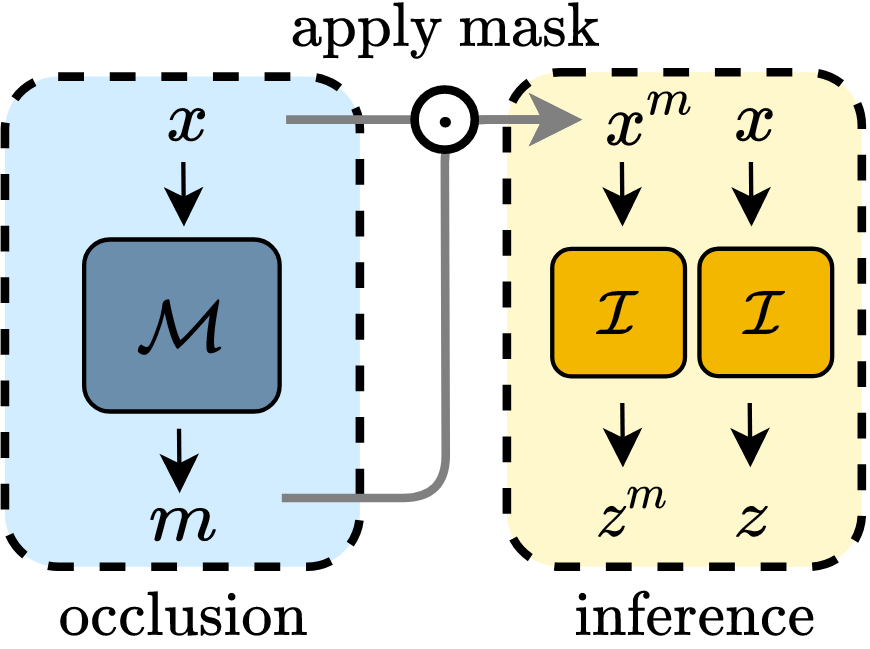}
  \vspace{-0.5\baselineskip}
  \caption{\!\adios\ Architecture.}
  \label{fig:architecture}
\end{wrapfigure}

\paragraph{Set up}%
ADIOS consists of two components, inference model $\I$ and occlusion model $\M$ (see \Cref{fig:architecture}). 
Given an RGB image $\x$, the occlusion model produces an image-sized mask $\m=\M(\x)$ with values in $[0, 1]$.
%
The inference model $\I$ takes original image $\x$ and an occluded image $\x^{\m} = \x\odot \m$
($\odot$ is the Hadamard product) as inputs, generating representations for both, which we denote as $\z$ and $\z^m$.
The two models are learnt by solving for
\begin{align}
    \I^\star, \M^\star = \arg\min_\I\max_\M \mathcal{L}(\x; \I, \M)\,. \label{eq:minmax_obj} 
\end{align}
We will now discuss different choices for $\I$ and $\M$.

\subsection{Inference model $\I$}
%
\setlength{\columnsep}{2ex}%
%
%
As discussed in \Cref{sec:intro}, the inference model should minimise some distance between the original and masked images.
Here, we discuss potential forms of this objective, arriving at our final framework using augmentation-based \gls{SSL} methods.
\paragraph{Distance in pixel space} \label{sec:ae_imputer}
\begin{wrapfigure}[8]{r}{0.3\linewidth}
    \vspace{-1.5em}
    \centering
    \includegraphics[width=0.77\linewidth]{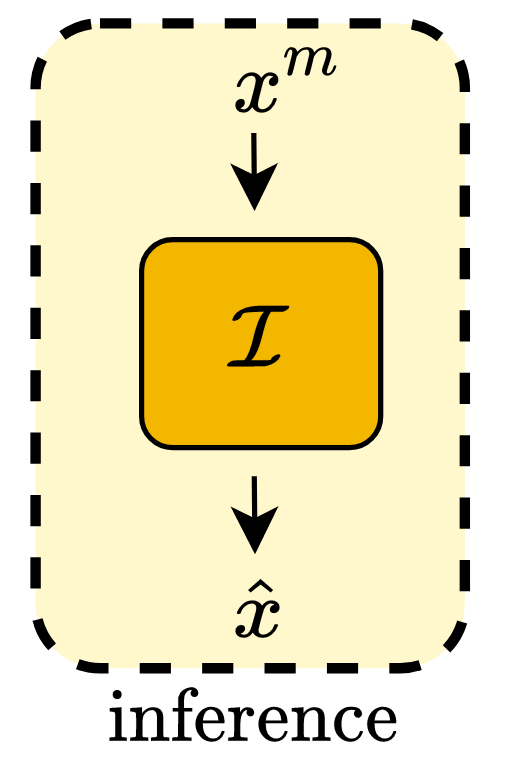}
    \vspace{-1em}
    \caption{Inpainting.}
    \label{fig:architecture_inpaint}
\end{wrapfigure}
One option would be to inpaint the masked image with the inference model, and train $\I$ by minimising the distance between the inpainted image and the original image in pixel space.
More specifically, we can define $\I$ as an auto-encoder consisting of an encoder and decoder, which takes the masked image $\x^m$ as input and produces inpainted image $\hat{x}$ (see \Cref{fig:architecture_inpaint}).
The model can be trained using the following reconstruction loss
\begin{align}
  \mathcal{L}_{\text{AE}}(\x; \I, \M)
  \!=\! \mathcal{D}(\x, \hat{\x})
  \!=\! \mathcal{D}(\x, \I (\x\odot \M(\x))), \label{eq:imputer_ae_obj}
\end{align}
where $\mathcal{D}$ denotes some distance metric defined in pixel space.
Minimising \cref{eq:imputer_ae_obj} encourages the auto-encoder to impute the missing part of the image as accurately as possible. 
$\mathcal{L}_{\text{AE}}$ can then be used in \cref{eq:minmax_obj} to train the inference-occlusion model.

\paragraph{Distance in representation space}
An interesting question for auto-encoding $\I$ is: where does the imputation happen? 
Multiple hypotheses exist: 
\circled{1} \emph{\textbf{Encoder only:}} The encoder $q$ completely recovers the missing
information in the masked image, and in the ideal case, $q(\M(\x)) = q(\x)$; the decoder
faithfully reconstructs these representations.
\circled{2} \emph{\textbf{Decoder only:}} The encoder faithfully extracts all information from the masked image, and the decoder reasons to inpaint missing pixel from the representations;
\circled{3} \emph{\textbf{Both:}} The encoder and decoder both inpaint parts of the image.

Given these scenarios, \circled{1} is clearly best suited for representation learning, as it requires the encoder to reason about the missing parts based on observed context, beyond just extracting image features.
With representation learning, rather than inpainting, being our end goal, the key challenge lies
in designing an objective targetting scenario \circled{1}, such that we learn the most expressive version of encoder $q$.

A key feature in \circled{1} is that when the encoder recovers all information of the original image, $q(\x^{\m}) = q(\x)$, the features extracted from the partially observed image $\x^{\m}$ should in principle be the same as those extracted from the unmasked image $\x$.
We thus propose an inference model $\I$ consisting of only the encoder, which extracts
representation $\z =\I(\x),\, \z\in\RR^d$.
Our objective can thus be written as 
\begin{align}
  \hspace*{-10pt}\mathcal{L}_{\text{ENC}}(\x;\I,\M)
  \!=\!\mathcal{D}(\z,\z^{\m}) 
  \!=\! \mathcal{D}(\I(\x), \I(\x\!\odot\! \M(\x)))
  \hspace*{-5pt}\label{eq:imputer_siamese_obj}
\end{align}
where $\mathcal{D}$ is some distance metric defined in $\RR^d$.
Not only does $\loss_{\text{ENC}}$ encourages the learning of more expressive encoder that can infer missing information, optimising this objective also does not involve generative component $p$, which is redundant for representation learning.

\begin{wrapfigure}[9]{r}{0.58\linewidth}
  \vspace{-1\baselineskip}
  \includegraphics[width=\linewidth]{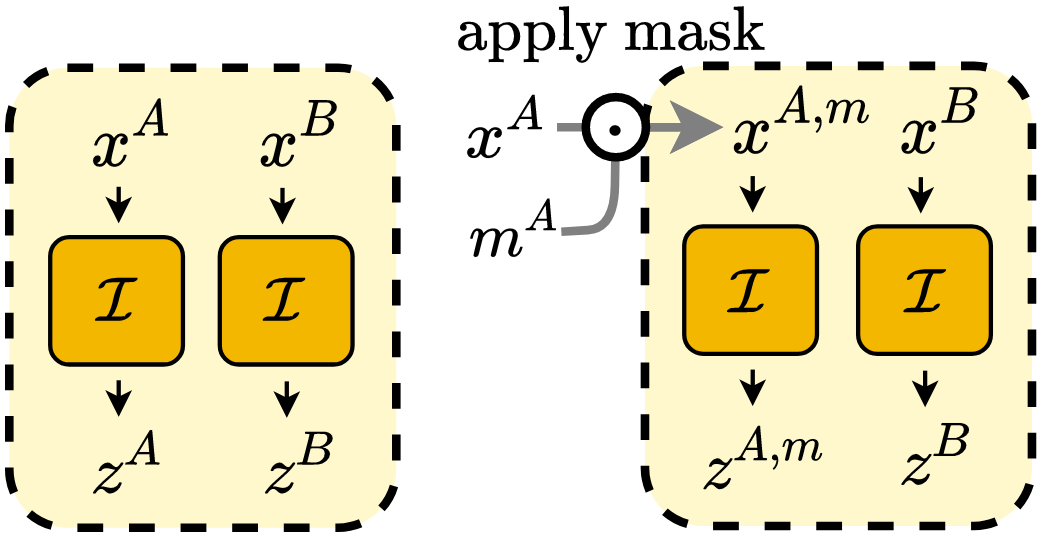}
  \vspace{-1.5\baselineskip}
  \caption{\textbf{\emph{Left}}:~SimCLR. \textbf{\emph{Right}}: SimCLR + ADIOS.}
  \label{fig:simclr+imputer}
\end{wrapfigure}
\paragraph{\acrshort{SSL} framework} 
\cref{eq:imputer_siamese_obj} can be realised by simply optimising a Siamese network \citep{bromley1993siamese}.
However, the objective can be trivially minimised when the representations for all inputs ``collapse'' to a constant.
This phenomenon, known as latent collapse, has been addressed in many ways in augmentation-based \acrshort{SSL}.
%
%

Let us take SimCLR \citep{chen2020simclr} as an example (see \Cref{fig:simclr+imputer}, left); given a minibatch of $M$ input images $\x =\{\x_i\}_{i=1}^{M}$, two sets of random augmentations $A$ and $B$ are applied to each image in $\x$, yielding $\x^A$ and $\x^B$. 
The same encoding function $\I$ is used to extract representations from both sets of augmented views, yielding $\z^A=\I(\x^A)$ and $\z^B=\I(\x^B)$. 
The objective of SimCLR is defined as
\begin{align}
    \mathcal{L}_{\text{SimCLR}}(\x; \I) =  \log \frac{\exp(\mathcal{D}(\z^A_i, \z^B_i))}{\sum_{i \neq j}\exp(\mathcal{D}(\z^A_i, \z^B_j))}\,, \label{eq:simclr_obj}
\end{align}
where $\mathcal{D}$ denotes the negative cosine similarity\footnote{We omit the SimCLR temperature parameter for simplicity.}. 
Intuitively, the objective minimises the distance between representations of the two augmented views of the same image (i.e. $\z^A_i, \z^B_i$), while repulsing the representations of different images (i.e. $\z^A_i, \z^B_j$).
This effectively prevents the representations of different images from collapsing to the same constant, while optimising an objective similar to \Cref{eq:imputer_siamese_obj}.
%

We can use the SimCLR objective for our model by masking one of the augmented images and then follow the exact same pipeline (see \Cref{fig:simclr+imputer}, right).
More specifically, we replace $\z^A$ by $\z^{A,\m}=\I(\x^A\odot \m^A)$, where $\m^A = \M(\x^A)$ is a mask generated by the occlusion model given $\x^A$.
Following \Cref{eq:simclr_obj}, we can write the SimCLR-ADIOS objective as
\begin{align}
  &\mathcal{L}_{\text{SimCLR}}^{\text{ADIOS}}(\x; \I,  \M)
    = \log \frac{\exp(\mathcal{D}(\z^{A,\m}_i, \z^B_i))}{\sum_{i \neq j}\exp(\mathcal{D}(\z^{A,\m}_i, \z^B_j))} \notag \\ 
  &= \log \frac{\exp(\mathcal{D}(\I(\x^A_i\odot\M(\x^A_i)), \I(\x^B_i)))}{\sum_{i \neq j}\exp(\mathcal{D}(\I(\x^A_i\odot\M(\x^A_i)), \I(\x^B_j)))}\,.
    \label{eq:simclr_adios_obj}
\end{align}
Again, we can use \cref{eq:simclr_adios_obj} in \Cref{eq:minmax_obj} to train the inference-occlusion model.
Crucially, any \acrshort{SSL} method that compares two augmented image views includes a term like \Cref{eq:imputer_siamese_obj}, and can be plugged in to our framework.
We conduct experiments using SimCLR, BYOL \cite{grill2020byol}, and SimSiam \cite{chen2021simsiam} objectives and show significant improvement on downstream task performance with each method.
Refer to \Cref{sec:app_objectives} for the ADIOS objective used for BYOL and SimSiam, as well as more details on the SimCLR objective.

\subsection{Occlusion model $\M$}
For simplicity, we only consider the single-mask-generating case in the discussion above.
In practice, since an image typically contains multiple components, we generate $N>1$ masks to challenge the model to reason about relations between different components---empirical performance confirm benefits of doing so.

There are many parametric forms $\M$ could employ.
For instance, one could consider generating multiple masks sequentially in an auto-regressive manner as seen in \citet{engelcke2021genesis}.
However, we find that the simplest setup suffices, where $\M: \RR^{c \times w \times h} \mapsto \RR^{N \times w \times h}$ consists of a learnable neural network and a pixelwise softmax layer $\sigma$ applied across $N$ masks to ensure that the sum of a given pixel across all the masks equals 1.
We use U-Net as the backbone of our occlusion model---see \Cref{sec:app_mask_arch} for more details.
Note that we experimented with binarising the masks during training, but found that this did not yield improvements, and hence used real-valued masks directly.

\subsection{Putting it together}

\begin{wrapfigure}[8]{r}{0.58\linewidth}
  \vspace{-2\baselineskip}
  \includegraphics[width=1.05\linewidth]{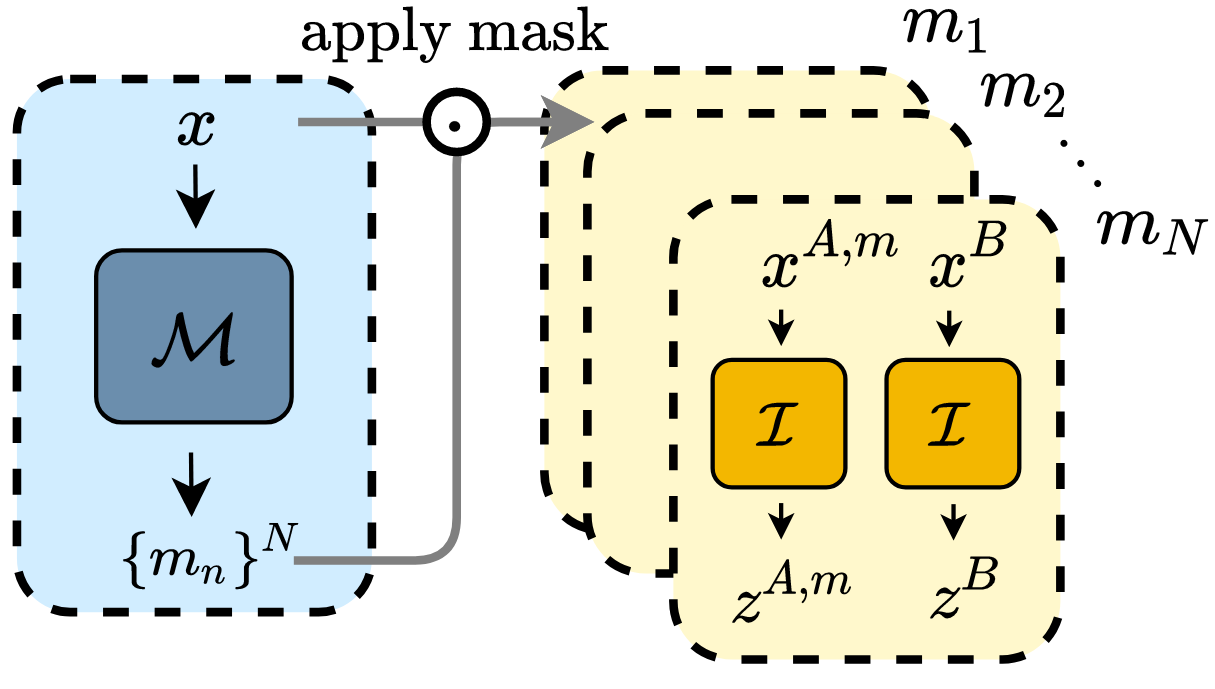}
  \vspace{-24pt}
  \caption{ADIOS, $N>1$.} 
  \label{fig:architecture_k}
\end{wrapfigure}

Here we present \adios\ in its complete form.
This includes the $N$-mask occlusion model, which generates masks $\{\m^{(n)}\}_{n=1}^N$ from the RGB image $\x$. 
The inference model computes a loss $\mathcal{L}^{(n)}(\x;\I,\M)$ for each $\m^{(n)}$ and the final loss is computed by averaging across $N$ masks
%
\begin{align}
    \I^\star, \M^\star = \arg\min_\I\max_\M  \frac{1}{N} \sum^N_{n=1} \mathcal{L}^{(n)}(\x; \I, \M)\,. \label{eq:minmax_obj_n}
\end{align}

\begin{wrapfigure}[9]{r}{0.32\linewidth}
  \centering
  \vspace*{-\baselineskip}
  \includegraphics[width=0.92\linewidth]{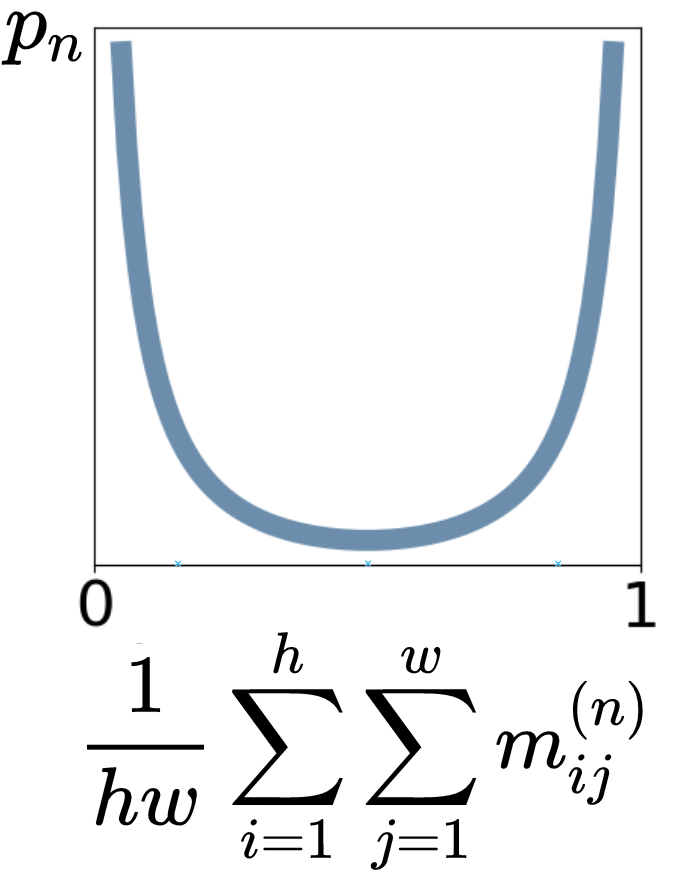}
  \vspace{-0.8\baselineskip}
  \caption{Penalty.}
  \label{fig:penalty_plot}
\end{wrapfigure}

\paragraph{Sparsity penalty}
A trivial solution exists for this objective \Cref{eq:minmax_obj_n}, where, for masks $\{\m^{(1)}, \ldots, \m^{(N)}\}$, some mask $\m^{(n)}$ occludes everything, with the other \scalebox{0.98}{$\{N\}_{\setminus n}$} masks not occluding anything. 
To avoid such degenerate solutions, we introduce a sparsity penalty $p_n$ in the form of $\nicefrac{1}{sin(\cdot)}$ that discourages the occlusion model from generating all-one or all-zero masks; specifically,
\begin{align}
    p_n = \sin\left(\frac{\pi}{hw}\displaystyle \sum_{i=1}^h \sum_{j=1}^w \m_{ij}^{(n)} \right)^{-1}\,.
\end{align}
Note that, $p_n$ goes to infinity as $\m^{(n)}$ approaches all-one or all-zero (see \Cref{fig:penalty_plot}).
Minimising $p_n$ with respect to $\M$ encourages the occlusion model to generate semantically meaningful mask, while avoiding degenerate solutions.

\paragraph{Final objective} Let $\lambda$ the scaling of the penalty term. Our complete objective reads as
\begin{align}
    \I^\star, \M^\star = \arg\min_\I\max_\M  \frac{1}{N} \sum^N_{n=1} \left( \mathcal{L}^{(n)}(\x; \I, \M) - \lambda p_n \right)\,. \label{eq:final}
\end{align}
\paragraph{Lightweight ADIOS}
Despite its strong empirical performance, we note that the training objective in \Cref{eq:final} requires $N$ forward passes, which can be computationally expensive as we increase $N$ for more complex data.
We therefore develop a lightweight version of ADIOS, where we randomly sample one from the $N$ generated masks to be applied to the input image.
Doing so disassociates the computational cost of the model from the number of generated masks, and the only cost increase comes from applying the mask generation model once, which is inexpensive ($10\%$ the size of ResNet18).
We name this single-forward pass version of our model ADIOS-s, and write the objective as
\begin{align}
    \I^\star, \M^\star =& \arg\min_\I\max_\M  \left( \mathcal{L}^{(k)} (\x; \I, \M) - \lambda \frac{1}{N} \sum^N_{n=1} p_n \right),\notag \\
    & \hspace{20pt} \text{where } k\sim \text{Uniform}\left(\{1,2...,N\}\right) . \label{eq:final_light}
\end{align}

%



%% file: 3experiments.tex
\section{Evaluation of Representations}
\label{sec:emp_eval}
\paragraph{Set up}
We evaluate \adios\ with three different \gls{SSL} objectives: SimCLR \citep{chen2020simclr}, BYOL \citep{grill2020byol}, and SimSiam \citep{chen2021simsiam}.
Each set of quantitative results is reported as an average over three random trials.
We summarise our training setup in \Cref{sec:app_hyper}.

\subsection{Classification}
We evaluate the performance of ADIOS on STL10, as well as a downsized version of ImageNet100 \citep{tian2020imagenet100}, from resolution 224x224 to 96x96.
We refer to our version of the dataset as ImageNet100-S.
We also evaluate the performance of ADIOS-s on the original ImageNet100 dataset.
Both ImageNet100 and STL10 are derived from ImageNet-1k \citep{imagenet}: ImageNet100 contains data from 100 ImageNet classes, and STL10 is derived from 10 object classes of ImageNet, with 5,000 labelled images and 100,000 unlabelled images.
Due to computational constraints we were unable to evaluate on ImageNet-1k; we leave this for future work.

For ADIOS, we provide results using ResNet18 \citep{resnet} and \vit-Tiny \citep{dosovitskiy2021vit} backbones on three classification tasks: linear evaluation, \knn~ and clustering. 
Through hyperparameter search, we use $N=4$ masking slots for ImageNet100-S and $N=6$ for STL10.
For ADIOS-s we provide results using ResNet18 as backbone on linear evaluation.

\paragraph{Linear evaluation and \knn}
We study the utility of learned representation by classifying the features using both linear and a $k$-nearest neighbour (\knn) classifiers.
Following the protocol in \citet{zhou2021ibot}, we sweep over different numbers of nearest neighbours for \knn~and different learning rates for the linear classifier.

\begin{table}[t]
\centering
\caption{Top-1 classification accuracy (\knn~and Linear Probing) on Imagenet100-S, STL10.  Improvements of ADIOS that are more than $3\%$ are marked in bold.}
\vspace*{0.5\baselineskip}
\scalebox{0.78}{
\begin{tabular}{lccccc}    \toprule
    \multirow{2}{*}{Method} & \multicolumn{2}{c}{ImageNet100-S} & & \multicolumn{2}{c}{STL10}  \\
    \cmidrule{2-3} \cmidrule{5-6}
    &  \knn & Linear & & \knn & Linear \\
    \midrule 
    \multicolumn{3}{l}{\textit{Backbone: \vit-Tiny}} &&& \vspace{2pt} \\ \rowcolor{lightblue}
    SimCLR       &  40.0 \std{0.28} & 40.2 \std{0.47} && 72.9 \std{0.27} & 76.0 \std{0.33}\\ \rowcolor{lightblue}
    \quad +ADIOS &  42.0 \std{1.32} & 43.1 \std{0.71} && 73.4 \std{0.28} & \textbf{79.7} \std{0.88}\\ 
    SimSiam      &  35.2 \std{1.12} & 36.8 \std{1.82} && 66.7 \std{0.10} & 67.5 \std{0.02}\\
    \quad +ADIOS &  \textbf{38.8} \std{2.73} & \textbf{40.1} \std{0.59} && 67.9 \std{0.75} & 68.8 \std{0.25}\\ \rowcolor{lightblue}
    BYOL         &  38.1 \std{0.61} & 39.7 \std{0.50} && 71.9 \std{0.12} & 72.1 \std{0.32}\\ \rowcolor{lightblue}
    \quad +ADIOS &  \textbf{47.1} \std{0.35} & \textbf{49.2} \std{0.94} && 74.5 \std{0.58} & \textbf{75.9} \std{0.63}\\ \midrule 
    \multicolumn{3}{l}{\textit{Backbone: ResNet-18}} &&& \vspace{2pt} \\ \rowcolor{lightblue}
    SimCLR       &  54.1 \std{0.09} & 55.1 \std{0.15} && 83.7 \std{0.48} & 85.1 \std{0.12}\\ \rowcolor{lightblue}
    \quad +ADIOS &  55.1 \std{0.43} & 55.9 \std{0.21} && 85.8 \std{0.10} & 86.1 \std{0.36}\\ 
    SimSiam      &  58.6 \std{0.31} & 59.5 \std{0.31} && 84.3 \std{0.81} & 84.8 \std{0.72}\\ 
    \quad +ADIOS &  61.0 \std{0.29} & 60.4 \std{0.19} && 84.6 \std{0.35} & 86.4 \std{0.24}\\ \rowcolor{lightblue}
    BYOL         &  56.2 \std{0.79} & 56.3 \std{0.10} && 83.6 \std{0.09} & 84.3 \std{0.13}\\ \rowcolor{lightblue}
    \quad +ADIOS &  \textbf{60.2} \std{0.82} & \textbf{61.4} \std{0.14} && 84.8 \std{0.19} & 85.6 \std{0.24}\\
    \bottomrule
\end{tabular}
\label{tab:knn-linear}}
\vspace*{-\baselineskip}
\end{table}

Results are presented in \Cref{tab:knn-linear}, where each \textit{+ADIOS} entry represents the ADIOS framework applied to the SSL objective in the row above.
For instance, the top coloured block shows results of SimCLR and SimCLR+ADIOS.
Models using ADIOS consistently outperform their respective SSL baselines beyond the margin of error; in some cases achieving significant improvements of 3--9$\%$ (in bold).
%
%
Notably, the \vit-Tiny models perform significantly worse than ResNet-18 models, which is unsurprising given that \vit-Tiny uses half the number of parameters of ResNet-18.

The best Top-1 accuracy on ImageNet100-S is 61.4$\%$, achieved by BYOL+ADIOS using linear evaluation, surpassing its baseline BYOL by 5.1$\%$; 
For STL10, the best performing model is SimSiam+ADIOS using linear evaluation with an accuracy of 86.4$\%$, while SimSiam evaluates at 84.8$\%$.
Significantly, \adios\ improves all metrics on both backbones and both datasets.
Interestingly, the degree of improvement varies by method, and can result in order change between respective \gls{SSL} methods.

We also run experiments on the original ImageNet100 dataset with the single-forward pass version of our model, ADIOS-s. 
The results in \Cref{tab:imagenet100_linear} show that this much cheaper model also achieves impressive performance, especially when applied to BYOL with a performance boost of more than $6\%$.
This result further demonstrates the efficiency of our approach, and the reduced computational cost allows for the potential of scaling to larger datasets.

{
\setlength{\aboverulesep}{0pt}
\setlength{\belowrulesep}{0pt}
\setlength{\extrarowheight}{.75ex}
\setlength\heavyrulewidth{0.3ex}
\begin{table}[t]
\vspace*{0.5\baselineskip}
\caption{Top-1 classification accuracy of linear probing on ImageNet100. Improvements of more than $3\%$ are marked in bold.}\label{tab:imagenet100_linear}
\vspace*{0.5\baselineskip}
\centering
\scalebox{0.7}{
\begin{tabular}{>{\columncolor{white}}c>{\columncolor{lightblue}}c|>{\columncolor{white}}c>{\columncolor{lightblue}}c|>{\columncolor{white}}c>{\columncolor{lightblue}}c}    \toprule
    SimCLR & +ADIOS-s & SimSiam & +ADIOS-s  & BYOL & +ADIOS-s  \\ \midrule
    77.5 \std{0.10} & 76.1 \std{0.50}  & 76.4 \std{0.07} & 77.2 \std{0.09} & 74.3 \std{0.16} & \textbf{80.8 \std{0.60}} \\
    \bottomrule
\end{tabular}}
\end{table}
}

\paragraph{Clustering}
Following \citet{bao2021beit,zhou2021ibot}, we also evaluate the trained models using standard clustering metrics, including adjusted random index (ARI) and Fowlkes-Mallows index (FMI), both of which computes the similarity between clusterings, as well as normalised mutual information (NMI).
We assign pseudo-labels to the representation of each image using k-means, and evaluate the three metrics on the clusters formed by the pseudo-labels vs. the true labels.
Results in \Cref{tab:unsupervised} are consistent with our previous findings.
ADIOS improves the performance of baseline SSL methods on all three metrics for both datasets.

\paragraph{Findings}
ADIOS significantly and consistently improves the quality of representation learned under a range of set-ups, across two datasets, two backbone architectures, three SSL methods and on five different metrics, highlighting the effectiveness and versatility of our approach.

\begin{table}[h]
\centering
\caption{Clustering performance on Imagenet100-S, STL10.}
\vspace*{0.5\baselineskip}
\scalebox{0.78}{
\begin{tabular}{lcccc}    \toprule
    \multirow{2}{*}{Method} &\multirow{2}{*}{Backbone} &  \multicolumn{3}{c}{Metrics}   \\
    \cmidrule{3-5} 
        &     & FMI \(\uparrow\) & ARI \(\uparrow\) & NMI \(\uparrow\) \\
    \midrule
    \multicolumn{3}{l}{\textit{Dataset: ImageNet100-S}} && \vspace{2pt} \\ \rowcolor{lightblue}
    SimCLR       & \vit-Tiny & 0.105 \std{\sci{1}{3}} & 0.095  \std{\sci{1}{3}}& 0.432 \std{\sci{3}{3}} \\ \rowcolor{lightblue}
    \quad +ADIOS & \vit-Tiny & 0.120 \std{\sci{1}{3}} & 0.110 \std{\sci{1}{3}} & 0.442 \std{\sci{4}{3}} \\
    SimSiam      & \vit-Tiny & 0.077 \std{\sci{9}{4}} & 0.067 \std{\sci{2}{3}} & 0.389 \std{\sci{3}{3}} \\
    \quad +ADIOS & \vit-Tiny & 0.098 \std{\sci{1}{2}} & 0.087 \std{\sci{9}{4}} & 0.425 \std{\sci{3}{3}} \\ \rowcolor{lightblue}
    BYOL         & \vit-Tiny & 0.098 \std{\sci{8}{3}} & 0.088 \std{\sci{8}{3}} & 0.418 \std{\sci{4}{3}} \\ \rowcolor{lightblue}
    \quad +ADIOS & \vit-Tiny & 0.132 \std{\sci{3}{3}} & 0.123 \std{\sci{1}{3}} & 0.458 \std{\sci{4}{3}} \\ 
    SimCLR       & ResNet18 & 0.151 \std{\sci{3}{3}}  & 0.135 \std{\sci{4}{3}} & 0.515 \std{\sci{6}{3}} \\
    \quad +ADIOS & ResNet18 & 0.175 \std{\sci{1}{3}}  & 0.161 \std{\sci{4}{3}} & 0.539 \std{\sci{3}{3}} \\ \rowcolor{lightblue}
    SimSiam      & ResNet18 & 0.167 \std{\sci{2}{3}}  & 0.136 \std{\sci{6}{3}} & 0.553 \std{\sci{8}{3}} \\ \rowcolor{lightblue}
    \quad +ADIOS & ResNet18 & 0.179 \std{\sci{1}{3}}  & 0.161 \std{\sci{1}{3}} & 0.553 \std{\sci{1}{3}} \\
    BYOL         & ResNet18 & 0.170 \std{\sci{1}{3}}  & 0.158 \std{\sci{3}{3}} & 0.530 \std{\sci{4}{3}} \\
    \quad +ADIOS & ResNet18 & 0.179 \std{\sci{6}{4}}  & 0.156 \std{\sci{2}{3}} & 0.561 \std{\sci{2}{3}} \\ \midrule
    \multicolumn{3}{l}{\textit{Dataset: STL10}} && \vspace{2pt} \\ \rowcolor{lightblue}
    SimCLR       & \vit-Tiny &0.349 \std{\sci{5}{3}} & 0.269 \std{\sci{6}{3}} & 0.410 \std{\sci{2}{3}} \\ \rowcolor{lightblue}
    \quad +ADIOS & \vit-Tiny &0.351 \std{\sci{4}{3}} & 0.271 \std{\sci{8}{3}} & 0.417 \std{\sci{6}{3}} \\
    SimSiam      & \vit-Tiny &0.296 \std{\sci{3}{3}} & 0.177 \std{\sci{1}{3}} & 0.341 \std{\sci{4}{3}} \\
    \quad +ADIOS & \vit-Tiny &0.320 \std{\sci{3}{3}} & 0.235 \std{\sci{5}{3}} & 0.349 \std{\sci{0}{0}} \\ \rowcolor{lightblue}
    BYOL         & \vit-Tiny &0.349 \std{\sci{5}{3}} & 0.269 \std{\sci{5}{3}} & 0.410 \std{\sci{5}{3}} \\ \rowcolor{lightblue}
    \quad +ADIOS & \vit-Tiny &0.355 \std{\sci{4}{2}} & 0.276 \std{\sci{3}{3}} & 0.422 \std{\sci{4}{3}} \\ 
    SimCLR       & ResNet18  &0.338 \std{\sci{2}{3}} & 0.166 \std{\sci{9}{4}} & 0.512 \std{\sci{5}{3}} \\
    \quad +ADIOS & ResNet18  &0.437 \std{\sci{6}{3}} & 0.309 \std{\sci{9}{3}} & 0.585 \std{\sci{8}{3}} \\ \rowcolor{lightblue}
    SimSiam      & ResNet18  &0.392 \std{\sci{2}{3}} & 0.242 \std{\sci{7}{3}} & 0.552 \std{\sci{3}{3}} \\ \rowcolor{lightblue}
    \quad +ADIOS & ResNet18  &0.412 \std{\sci{8}{3}} & 0.249 \std{\sci{7}{3}} & 0.558 \std{\sci{2}{4}} \\
    BYOL         & ResNet18  &0.429 \std{\sci{5}{3}} & 0.328 \std{\sci{9}{3}} & 0.525 \std{\sci{8}{3}} \\
    \quad +ADIOS & ResNet18  &0.508 \std{\sci{6}{3}} & 0.422 \std{\sci{1}{2}} & 0.588 \std{\sci{9}{3}} \\
    \bottomrule
\end{tabular}
\label{tab:unsupervised}}
\vspace*{-\baselineskip}
\end{table}

\subsection{Transfer learning}

We study the downstream performance of models trained on ImageNet100-S, on four different datasets including CIFAR10, CIFAR100 \citep{cifar}, Flowers102 \citep{flowers}, and iNaturalist \citep{inaturalist}.
CIFAR10 and CIFAR100 resolutions are 32, while those of Flowers102 and iNaturalist are 96.
We only use the ResNet-18 models here as they clearly outperform the \vit-Tiny models in our previous experiments.
Detailed setup and hyperparameters are given in \Cref{sec:app_hyper_transfer}.

\Cref{tab:transfer} reports classification accuracy under two different transfer-learning setups, including \textit{F.T}: fine-tuning the entire model and \textit{Lin.}: freeze the encoder weights and re-train the linear classifier only.
As a comparison, we also show the results of training from scratch on each dataset.

Results show that ADIOS improves transfer learning performance on all four datasets, under both linear evaluation and fine-tuning.
Bigger improvements, of $>$3$\%$ (marked in bold) occur mostly under linear evaluation, indicating that compared to baseline \gls{SSL} models, the \adios-pre-trained representations are much easier to linearly separate.
Notably, all 6 models' fine-tuning performance exceeds training from scratch, demonstrating the benefits of pretraining. 

One might also notice that different from the other datasets, there exists large discrepancies between the linear evaluation vs. fine-tuning performance on CIFAR10 and CIFAR100.
This is because \citet{resnet} suggests a slightly different architecture for CIFAR with smaller kernel size in the first layer to suit its small image size (see details in \Cref{sec:cifar_resnet}).
We use CIFAR-ResNet for fine-tuning, however we have to use the original architecture for linear evaluation in order to use the pretrained weights, which leads to poor performance.

\begin{table}[t]
\centering
\caption{Classification accuracy of transfer learning by re-training linear classifier only (\textit{Lin.}) and fine-tuning (\textit{F.T}). More than $3\%$ improvements by ADIOS are marked in bold.}
\vspace*{0.5\baselineskip}
\scalebox{0.78}{
\begin{tabular}{lcccccccc}    \toprule
    \multirow{2}{*}{Method} & \multicolumn{2}{c}{CIFAR10} & \multicolumn{2}{c}{CIFAR100} & \multicolumn{2}{c}{Flowers102}& \multicolumn{2}{c}{iNaturalist}  \\
    \cmidrule(lr){2-3} \cmidrule(lr){4-5} \cmidrule(lr){6-7}  \cmidrule(lr){8-9}
    & Lin. & F.T. & Lin. & F.T.  & Lin. & F.T.  & Lin. & F.T. \\
    \midrule \rowcolor{lightblue}
    SimCLR        & 30.1 & 91.3  & 10.2 & 70.0 & 42.5 & 45.6& 69.4 & 82.1 \\ \rowcolor{lightblue}
    \quad +ADIOS  & \textbf{34.6} & 93.4  & 11.0 & 71.8 & \textbf{50.2} & \textbf{50.6}& \textbf{72.5} & 84.3 \\ 
    SimSiam       & 35.3 & 92.4  & 13.2 & 65.0 & 38.7 & 55.0& 72.3 & 85.0 \\ 
    \quad +ADIOS  & \textbf{39.3} & 94.3  & 13.3 & \textbf{71.0} & \textbf{44.9} & \textbf{59.0} & \textbf{75.9} & 86.2\\  \rowcolor{lightblue}
    BYOL          & 29.9 & 88.0  & 13.3 & 52.3 & 49.1 & 58.6& 72.7 & 85.1 \\ \rowcolor{lightblue}
    \quad +ADIOS  & \textbf{39.2} & 90.4  & 14.0 & \textbf{62.0} & 51.7 & 60.1& 73.1 & 85.7 \\  \midrule
    {Scratch}& {-} & {85.5} & {-}    & {49.8} & {-} & {30.6} & {-} & {73.8} \\ 
    \bottomrule
\end{tabular}
\label{tab:transfer}}
\vspace*{-\baselineskip}
\end{table}

\subsection{Robustness}\label{sec:robustness}

\begin{wrapfigure}[12]{r}{0.4\linewidth}
  \vspace{-1em}
  \centering
  \hspace*{-2ex}
  \includegraphics[width=\linewidth]{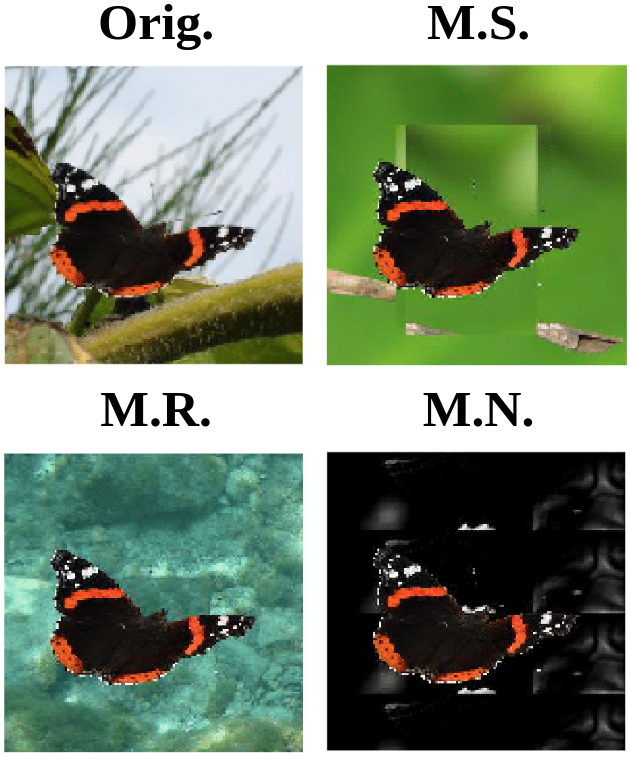}
  \vspace{-1ex}
  \caption{Examples.}
  \label{fig:bgchallenge}
\end{wrapfigure}

It is likely that the adversarial masks learned by \adios\ target informative image features, including spurious correlations if present in a dataset.
It is therefore interesting to ask if \adios\ representations are robust to changes in such spurious correlations.
To answer this, we evaluate models pretrained on ImageNet100-S on the backgrounds challenge \citep{bgchallenge}, where 7 different types of variation on a subset of ImageNet data are used to measure the impact of foreground and background on model decision making.
Examples of such variations can be seen in \Cref{fig:bgchallenge}, where the original figure's (\textit{Orig.}) background is replaced by background from another image in the same class (\textit{M.S.}), from a random image in any class (\textit{M.R.}) or from an image in the next class (\textit{N.R.}).
We perform linear evaluation on the pretrained models on these variations, and report the classification accuracy in \Cref{tab:bgchallenge}.

Our results show that all three SSL-ADIOS models outperform their respective baselines on \emph{all} variations, demonstrating that \adios-learned representations are more robust to changes in both foreground and background.

It is useful to examine how \adios\ behaves in different testing conditions regardless of the \gls{SSL} objective used.
To this end, the bottom row of \Cref{tab:bgchallenge} contains  the performance gain of \adios\ averaged over all three SSL models.
We witness the biggest gains on the \textit{M.R.} and \textit{M.N.} conditions (bottom row, \Cref{fig:bgchallenge}).
That is, when any deterministic relation between labels and backgrounds is severed.
The improvement is the lowest for the original images and \textit{M.S.} condition (top row, \Cref{fig:bgchallenge}), both of which preserve the relation between labels and background.
This demonstrates that \adios\ depends less on background information that are spuriously-correlated with object labels when making predictions.
This is not surprising---as we describe in \Cref{sec:qualitative}---\adios' generated masks tend to occlude backgrounds, encouraging the model to focus on the foreground objects.
%

\begin{table}[t]
\caption{Accuracy on different variations of the backgrounds challenge, evaluating model robustness. Example variations in \Cref{fig:bgchallenge}.}\label{tab:bgchallenge}
\vspace*{0.5\baselineskip}
\centering
\scalebox{0.72}{
\begin{tabular}{lcccccccc}    \toprule
    \multirow{2}{*}{Method} & \multicolumn{7}{c}{Variations} & \multirow{2}{*}{Orig.} \\
    \cmidrule{2-8}
    &  O.BB. & O.BT. & N.F. & O.F. & M.S.& M.R. & M.N. & \\
    \midrule \rowcolor{lightblue}
    SimCLR        & 20.1 & 34.8 & 44.3 & 41.6 & 67.1 & 45.9 & 41.0 & 78.8 \\ \rowcolor{lightblue}
   \quad +ADIOS   & 20.7 & 36.7 & 45.5 & 43.5 & 68.0 & 47.9 & 43.7 & 79.1 \\ 
    SimSiam       & 29.5 & 39.1 & 43.8 & 52.1 & 69.9 & 43.9 & 40.8 & 78.4 \\ 
    \quad +ADIOS  & 33.1 & 41.0 & 46.2 & 54.7 & 71.5 & 47.2 & 43.5 & 80.3 \\ \rowcolor{lightblue}
    BYOL          & 25.9 & 38.4 & 46.0 & 51.6 & 71.3 & 45.6 & 42.7 & 79.8 \\ \rowcolor{lightblue}
    \quad +ADIOS  & 27.7 & 39.0 & 48.5 & 51.7 & 72.1 & 47.8 & 44.1 & 80.6 \\ \midrule
    Avg. gain    & {+2.0} & {+1.4} & {+2.0} & {+1.5} & {+1.1} & {+2.5} & {+2.3} & {+1.0} \\
    \bottomrule
\end{tabular}}
\vspace*{-1.5\baselineskip}
\end{table}

\section{Analysis on Learned Masks}\label{sec:emp_analysis}


\begin{figure*}[t]
  \centering
  \captionsetup[subfigure]{belowskip=1ex}
  \begin{subfigure}{\linewidth}
    \centering
    \includegraphics[width=\linewidth]{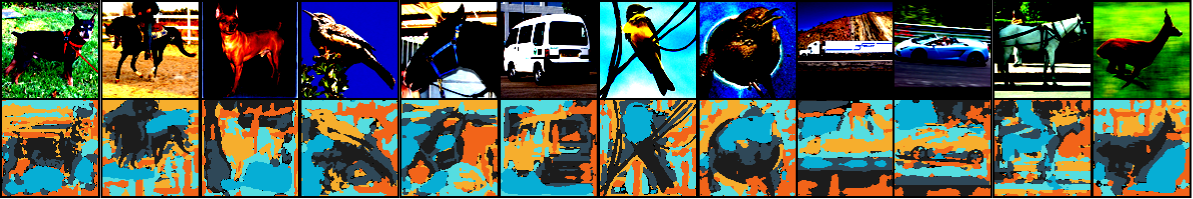}
    \caption{STL10, $N=6$.}\label{fig:stl10_mg}
  \end{subfigure}
  \begin{subfigure}{\linewidth}
    \centering
    \includegraphics[width=\linewidth]{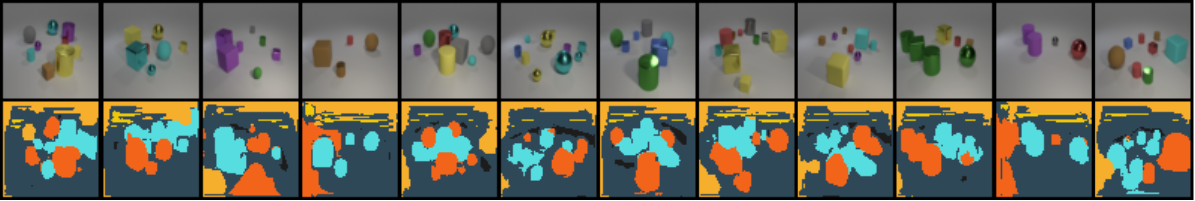}
    \caption{CLEVR, $N=4$.}\label{fig:clevr_mg}
  \end{subfigure} 
  \begin{subfigure}{\linewidth}
    \centering
    \includegraphics[width=\linewidth]{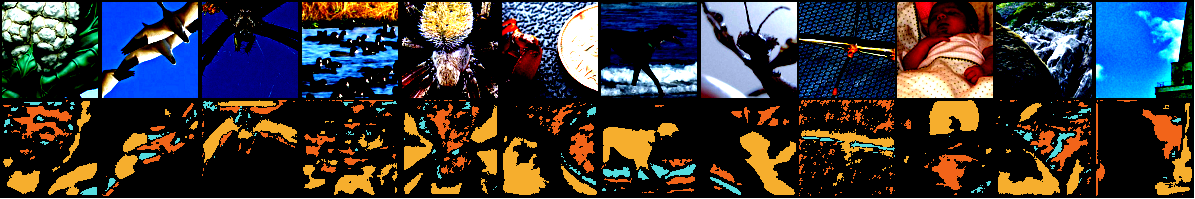}
    \caption{ImageNet100-S, $N=4$.}\label{fig:imagenet_mg}
  \end{subfigure}
  \vspace{-1em}
  \caption{Masks generated by ADIOS during training on each dataset. $N$ denotes the number of masks. \textit{Top row:} original image; \textit{Bottom row:} generated masks, each color represents one mask.}\label{fig:mask_gen}
\end{figure*}

Here, we look at the masks generated by \adios' occlusion model when trained on ImageNet100-S, STL10, and CLEVR \citep{clevr}---a dataset of rendered 3D objects such as cubes and spheres.
We use $N=4$ masking slots for CLEVR and include training details in \Cref{sec:app_hyper_clevr}.
The top row of each image block in \Cref{fig:mask_gen} shows the original image.
The bottom row displays the generated masks, with each colour representing one masking slot.
See \Cref{sec:qualitative} for a detailed analysis.
We also quantitatively analyse \adios' masks in \Cref{sec:mask_quant}.

\subsection{Mask Generation} \label{sec:qualitative}
For realistic, single-object datasets such as STL10 and ImageNet100-S, ADIOS manages to mask out different compositions of the image.
In the STL10 dataset, each image clearly shows a `foreground' object and a `background' (\Cref{fig:stl10_mg}).
In this setting, \adios\ learns to mask specific object parts like the wings or the tail of a bird (4$^\text{th}$ column) and the mouth or the face of a horse (5$^\text{th}$ column).
In case of the ImageNet100-S dataset, however, it is often not obvious if the image features any particular entity.
Hence, \adios\ tends to occlude complete entities like the animal and the ant in the in the 7$^\text{th}$ and 8$^\text{th}$ columns of \Cref{fig:imagenet_mg}.

In CLEVR (simple rendered objects), \adios\ is usually able to put the background into a separate slot; the remaining slots split all present objects into 2-3 groups, with a tendency of applying a single mask to objects of the same colour (see \Cref{fig:clevr_mg}).
While not a focus on our work, and in contrast to prior art \cite{greff2019iodine,engelcke2019genesis}, \adios\ does not produce perfect segmentations.
It is however an interesting research direction, given that better segmentations could further improve representation learning performance as we show in \Cref{sec:mask_quant}.

\paragraph{Summary} Qualitative results show that ADIOS can generate semantically-meaningful masks.
Crucially, the generated masks focus on different levels of detail, depending on the dataset.
This may explain some of the \adios' performance gains in the robustness experiments of \Cref{sec:robustness}, as semantic perturbations are baked into the training process.

\subsection{Comparing Masking Schemes}
\label{sec:mask_quant}

The premise of our work is that for \acrlong{MIM}, \emph{what is masked}, matters.
In this section, we further investigate this by comparing the representation learning performance of SSL models trained under an \adios-like MIM framework, but with non-parametric masks including ground-truth semantic masks and random masks.

\begin{figure}[t]
  \centering
  \captionsetup[subfigure]{skip=2pt,belowskip=1ex}
  \begin{subfigure}{\linewidth}
    \centering
    \includegraphics[width=\linewidth]{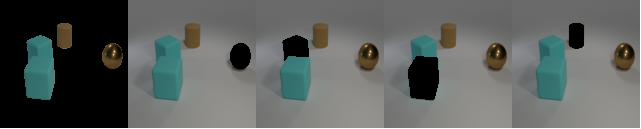}
    \caption{Ground-truth object masks}\label{fig:ms_gt}
  \end{subfigure}
  \begin{subfigure}{\linewidth}
    \centering
    \includegraphics[width=0.4\linewidth]{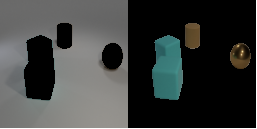}
    \caption{Foreground-background masks}\label{fig:ms_fgbg}
  \end{subfigure}
  \begin{subfigure}{\linewidth}
    \centering
    \includegraphics[width=\linewidth]{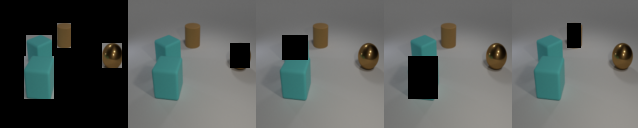}
    \caption{Ground-truth box-shaped masks}\label{fig:ms_box}
  \end{subfigure}
  \begin{subfigure}{\linewidth}
    \centering
    \includegraphics[width=\linewidth]{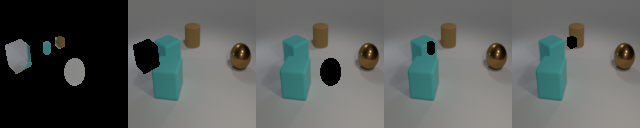}
    \caption{Shuffled ground-truth masks}\label{fig:ms_shuffle}
  \end{subfigure}
  \begin{subfigure}{\linewidth}
    \centering
    \includegraphics[width=\linewidth]{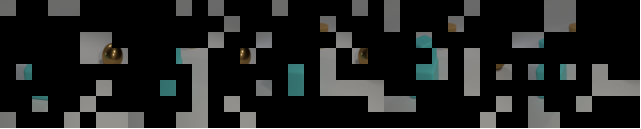}
    \caption{MAE \citep{he2021mae} masks}\label{fig:ms_mae}
  \end{subfigure}
  \begin{subfigure}{\linewidth}
    \centering
    \includegraphics[width=\linewidth]{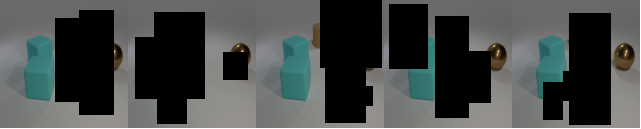}
    \caption{BEiT \citep{bao2021beit} masks}\label{fig:ms_beit}
  \end{subfigure}
  \caption{Masking schemes used to compare against ADIOS.}\label{fig:masking_scheme}
  \vspace*{-\baselineskip}
\end{figure}

In \Cref{fig:masking_scheme} we outline the different masking schemes investigated in this section on the CLEVR dataset, including:
\begin{inparaenum}[a)]
\item ground-truth object segmentation masks (provided with the dataset),
\item foreground-background masks (where the foreground is the union of all objects),
\item ground-truth, box-shaped masks,
\item shuffled ground-truth object segmentation masks (i.e. one image uses the ground-truth mask of another),
\item random mask occluding 75\% of the image, as in MAE \citep{he2021mae},
\item blockwise mask occluding 30\% of the image, same as BEiT \citep{bao2021beit}.
\end{inparaenum}
Note that out of these masking schemes, a--c include semantic information, while d--f do not.
In addition, we perform similar experiments on ImageNet100-S and STL10, however since we do not have information of the ground-truth object segmentation, we only compare the performance of ADIOS against using the random masking scheme in MAE and BEiT (i.e. e and f).


The representation learning performance of different masking schemes on ImageNet100-S and STL10 is evaluated by its top-1 classification accuracy under linear probing.
To evaluate this on CLEVR, we set up a challenging multi-label classification task.
Namely, we predict 24 binary labels, each indicating presence of a particular colour and shape ($8\times3$) combination in the image.
We report the F1 score (harmonic mean of the precision and recall) under different weighted average of subpopulation (defined by labels), where `micro' evaluates F1 across the entire dataset, `macro' evaluates an unweighted average of per-label F1 score, and `weighted' scale the per-label F1 score by number of examples when taking average.

%
\Cref{tab:clevr_quantitative,tab:im100stl10_quantitative} contains results averaged across these three \gls{SSL} methods, each with three random trials (i.e. each entry in \Cref{tab:clevr_quantitative,tab:im100stl10_quantitative} is averaged over nine runs).
This is to marginalize out particular \gls{SSL} methods, and therefore better understand effect of each of the masking schemes.

The results clearly show the advantage of using semantic masks over non-semantic masks for representation learning.
In \Cref{tab:im100stl10_quantitative}, we show that ADIOS significantly outperform random masking schemes used in MAE and BEiT;
additionally, the ground-truth object masks (\textit{G.T.}) In \Cref{tab:clevr_quantitative} achieves the best performance on all three metrics, closely followed by ADIOS with comparable F1-macro, F1-weighted score, and slightly lower F1-micro score.
We compare this to the randomly shuffled ground-truth mask (\textit{Shuffle}), covering on average the same image fraction as the ground-truth, but where there is far less semantic consistency to the image content.
The shuffled masks perform much worse on all three metrics, supporting our hypothesis from \Cref{sec:intro} that \emph{what is masked} is more important that \emph{how much is masked}.

\begin{table}[t]
\centering
\caption{Top-1 classification accuracy on ImageNet100-S and STL10 under different masking schemes, averaged over three runs of SimCLR, SimSiam and BYOL respectively. Best results for each metric in bold.}
\vspace*{0.5\baselineskip}
\scalebox{0.76}{
\begin{tabular}{clcc} \toprule
    \multirow{2}{*}{\shortstack{Mask type}} & \multirow{2}{*}{Condition} & \multicolumn{2}{c}{Dataset} \\ \cmidrule{3-4}
     & & ImageNet100-S & STL10  \\
    \midrule 
    \multirow{2}{*}{\shortstack{Random}}
    & e) MAE     &43.7  \std{0.43} & 78.4  \std{0.91}   \\
    & f) BEiT    &46.4  \std{0.67} & 80.7  \std{1.00}   \\ \midrule
    Learned & ADIOS& \textbf{59.2  \std{2.92}} & \textbf{86.0  \std{0.40}}   \\ \midrule
    None & -    & 57.0  \std{2.27} & 84.7  \std{0.40} \\
    \bottomrule
\end{tabular}\label{tab:im100stl10_quantitative}}
\vspace*{-\baselineskip}
\end{table}

\begin{table}[t]
\centering
\caption{Multi-label classification on CLEVR under different masking schemes, averaged over three runs of SimCLR, SimSiam and BYOL respectively. Best results for each metric in bold.}
\vspace*{0.5\baselineskip}
\scalebox{0.76}{
\begin{tabular}{clccc} \toprule
    \multirow{2}{*}{\shortstack{Mask type}} & \multirow{2}{*}{Condition} & \multicolumn{3}{c}{Metric} \\ \cmidrule{3-5}
     & & F1-macro \(\uparrow\) & F1-micro \(\uparrow\) & F1-weighted \(\uparrow\)\\
    \midrule 
    \multirow{3}{*}{\shortstack{Semantic}} & a) G.T. & \textbf{0.373 \std{7e-3}} & \textbf{0.401 \std{2e-4}} & \textbf{0.460 \std{1e-2}}  \\
    &b)  FG./BG.  & 0.346 \std{7e-3} & 0.365 \std{2e-4} & 0.402 \std{1e-3}  \\
    &c)  Box       & 0.347 \std{2e-4} & 0.391 \std{3e-5} & \textbf{0.457 \std{5e-2}}  \\ \midrule
    \multirow{3}{*}{\shortstack{Random}}
    & d) Shuffle      &0.332 \std{6e-3} & 0.360 \std{8e-4} & 0.418 \std{1e-3}  \\
    & e) MAE    &0.309 \std{8e-4} & 0.336 \std{3e-4} & 0.391 \std{9e-4}  \\
    & f) BEiT    &0.274 \std{1e-3} & 0.307 \std{2e-4} & 0.395 \std{7e-3}  \\ \midrule
    Learned & ADIOS& \textbf{0.377 \std{2e-3}} & 0.385 \std{9e-4} & \textbf{0.451 \std{1e-3}}  \\ \midrule
    None & -    & 0.352 \std{9e-3} & 0.359 \std{2e-4} & 0.373 \std{2e-5}  \\
    \bottomrule
\end{tabular}\label{tab:clevr_quantitative}}
\vspace*{-\baselineskip}
\end{table}

The remaining masking schemes behave as expected, with the semantically-informed ones outperforming the non-semantic ones.
Perhaps surprisingly, random masks, including the ones used in MAE and BEiT, can hurt representation learning under this \adios-like MIM framework: in both \Cref{tab:im100stl10_quantitative,tab:clevr_quantitative}, the performance of random masking schemes are much lower even compared to the baseline where no mask is applied.
We do note that MAE and BEiT both contain many components beyond their masking schemes that are critical for their successful learning as reported in the respective works.
These include image decoders, discrete-VAE tokeniser and the \vit  encoder.
Our evaluation focus exclusively on the masking schemes, and suggests that semantically meaningful masks lead to better representations, while random masks do not.

\paragraph{Summary}
Our experiments here show two things.
Firstly, that semantically meaningful masks can be used as an effective form of augmentation for SSL models, but the same cannot be said for random masks.
And secondly, that the representations learned from using the masks generated by ADIOS are comparable in quality to those learned from using ground-truth object masks.

%% file: 4relatedwork.tex
\section{Related Work}

\paragraph{Augmentation based \gls{SSL}}
Recent work has seen rapid development in \gls{SSL} utilising image augmentation, with the core idea being that the representation of two augmented views of the same image should be similar.
This involves a range of work that adopts a contrastive framework where the positive sample pairs (i.e. two views of the same image) are attracted and negative pairs (i.e. two views of different images) are repulsed \cite{chen2020simclr, gansbeke2020scan, he2020moco, chen2020mocov2, caron2020swav}, as well as non-contrastive approaches \cite{grill2020byol, ermolov2021wmse, zbontar2021barlow, chen2021simsiam, bardes2021vicreg} that are able to prevent latent collapse without negative pairs, which is considered to be computationally expensive.

\paragraph{Learning augmentations}
Several models have proposed to learn augmentation policies with supervision signals
\cite{cubuk2019autoaugment,hataya2020fasterautoaugment} that is more favourable to the task at hand.
\citet{tamkin2021viewmaker} applies these for \gls{SSL} and proposes to learn perturbation to the input image with a \emph{viewmaker} model that is trained adversarially to the main encoder network.
Different from our work where masks are generated to occlude different components of the image, their learned perturbation is $l_p$-bounded and provides more ``color-jitter'' style augmentation to the input. 
\citet{koyama2021contrastive} also proposes to learn mask-like augmentations by maximising a lower bound to the mutual information between image and representation while regularising the entropy of the augmentation.
However their experiment is limited to an edited MNIST dataset, and they only consider learning augmentations for one \gls{SSL} algorithm, SimCLR.

\paragraph{Masked image models}
More recently, models such as MAE \citep{he2021mae}, BEiT \citep{bao2021beit}, iBOT \citep{zhou2021ibot}
have been motivated by masked language models like BERT, as we ourselves do, and achieved highly competitive results on \gls{SSL}.
All three methods make use of vision transformers and propose to ``inpaint'' images occluded by \emph{random masks} in one way or another: 
MAE employs an autoencoder to inpaint images that are heavily masked, with the decoder discarded after pretraining.
On the other hand, BEiT and iBOT both utilise tokenisers to first transform image patches into visual tokens, with BEiT using off-the-shelf pretrained tokenizer and iBOT training the tokeniser online. 
Similar to our work, rather than performing reconstruction in pixel space, they minimise the distance between the visual tokens of the complete image vs. the masked image. 
Recent work has also seen the application of random masks to modalities beyond vision including speech and text, achieving strong performance in all these domains \citep{data2vec}.
Our work is significantly different from all above since we employ \emph{semantically meaningful masks} from the occluder model, jointly learned with the encoding model.
Moreover, our model does not rely on additional components such as tokenisers/image decoder, and since the construction of the model does not require splitting the image into patches, is also not limited to using vision transformers as the backbone architecture.

%% file: 5conclusion.tex
\section{Conclusion}\label{sec:conclusion}

We propose a novel MIM framework named ADIOS, which learns a masking function alongside an image encoder in an adversarial manner. 
We show, in extensive experiments, that our model consistently outperforms SSL baselines on representation learning tasks, while producing semantically-meaningful masks.
We also provide detailed analysis on using different forms of occlusion as augmentation for SSL in general.
We find that the best representation learning performance results from using semantically-meaningful masks, especially ground-truth ones, and that masks generated by the ADIOS' occlusion model are almost as good.

One caveat of our model is that the memory and computation cost increases linearly with the number of masking slots $N$, since the model requires $N$ forward passes before each gradient update.
This likely can be addressed by randomly sampling one mask for each forward pass, but is left to future work.
Additionally, we want to investigate ADIOS performance on larger datasets such as ImageNet-1K or 22K and larger backbones like \vit-L, \vit-H.
However, we believe that, as it is, ADIOS' strong performance on a variety of tasks under versatile conditions provides valuable insights on the design of masked image models.
Future work on \acrlong{MIM} should consider not only objectives and model architectures, but also mask design---this work shows that semantic masks are significantly more helpful than random ones in aiding representation learning.

%% file: 6acknowledgement.tex
\section{Acknowledgements}
We would like to thank Sjoerd van Steenkiste, Klaus Greff, Thomas Kipf, Matt Botvinick, Adam Goliński, Hyunjik Kim and Geoffrey E. Hinton for helpful discussions in the early stages of this project.
We would also like to thank Benjamin A. Stanley for discussions on the sparsity penalty term.

YS and PHST were supported by the UKRI grant: Turing AI Fellowship EP/W002981/1 and EPSRC/MURI grant: EP/N019474/1. We would also like to thank the Royal Academy of Engineering and FiveAI.
YS was additionally supported by Remarkdip through their PhD Scholarship Programme.

%% file: 6appendix1.tex
\clearpage
\appendix

\section{ADIOS Objectives} \label{sec:app_objectives}
In this section we detail the objective used to optimise ADIOS+SimCLR, ADIOS+SimSiam and ADIOS+BYOL.
\subsection{SimCLR}
\begin{wrapfigure}[11]{r}{0.55\linewidth}
  \vspace{-1.5\baselineskip}
  \includegraphics[width=\linewidth]{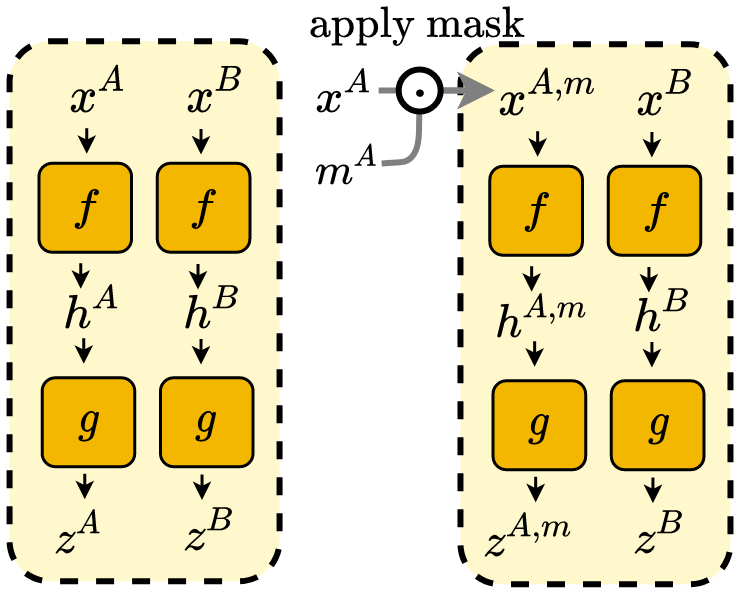}
  \vspace{-1.5\baselineskip}
  \caption{\textbf{\emph{Left}}:~SimCLR. \textbf{\emph{Right}}: SimCLR + ADIOS.}
  \label{fig:simclr_detailed}
\end{wrapfigure}
We show a simplified version of SimCLR architecture in \Cref{fig:simclr+imputer}.
In reality, the encoder $\I$ of SimCLR further factorises into two networks, including a base encoder $f(\cdot)$ which extracts the representation $\h$ used for downstream tasks, followed by a projection head $g(\cdot)$ which maps $\h$ to the final embedding that's used to compute the objective in \Cref{eq:simclr_obj}.

We visualise this in \Cref{fig:simclr_detailed}. 
It is helpful to establish SimCLR's architecture as it lays the foundation for both SimSiam and BYOL.

\subsection{SimSiam}
\begin{wrapfigure}[10]{r}{0.55\linewidth}
  \vspace{-1.5\baselineskip}
  \includegraphics[width=\linewidth]{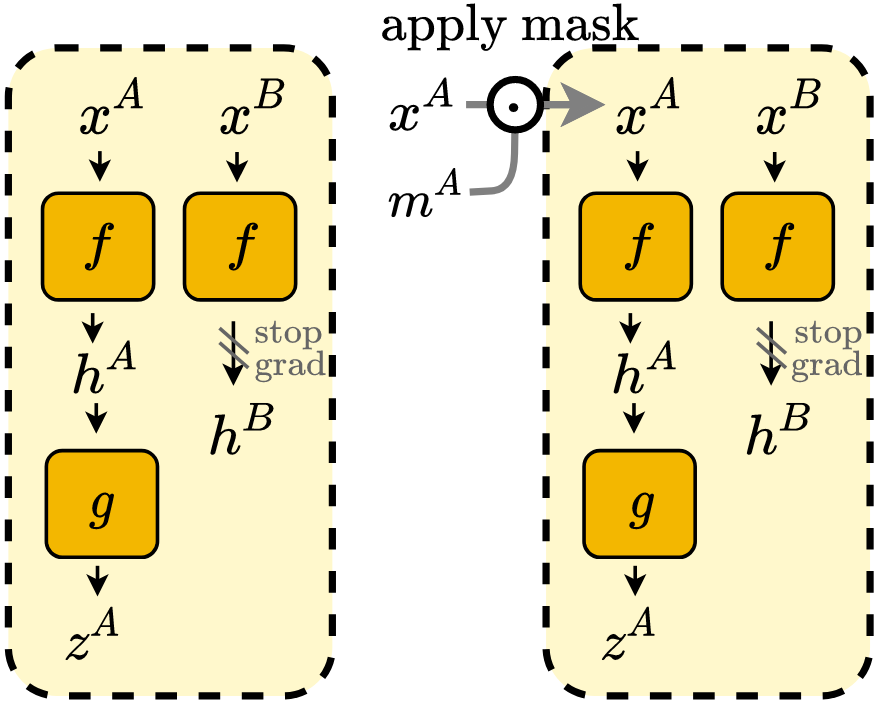}
  \vspace{-1.5\baselineskip}
  \caption{\textbf{\emph{Left}}:~SimSiam. \textbf{\emph{Right}}: SimSiam + ADIOS.}
  \label{fig:simsiam_detailed}
\end{wrapfigure}

\citet{chen2021simsiam} proposes SimSiam, an SSL method that can learn meaningful representations without negative examples.
The forward pass of SimSiam also contains a base encoder and a projection head; 
however, different from SimCLR, for one of the augmentation streams the projection head is removed with a stop gradient operation applied to the base encoder (See \Cref{fig:simsiam_detailed}).
Authors find empirically that these two alterations are essential for preventing latent collapse in the absence of negative examples.
The final objective of SimSiam is written as
\begin{align}
    \mathcal{L}_{\text{SimSiam}}(\x; \I) = \frac{1}{2}\left(\mathcal{D}(\z^A, \h^B) + \mathcal{D}(\z^B, \h^A) \right), \label{eq:simsiam}
\end{align}
where $\mathcal{D}$ denotes the negative cosine similarity,  $\I=g \circ f$, and $\z=g(f(\x))$ while $\h=f(\x)$.
Note that the loss is the average of two distances due to the asymmetrical model.

Following the same intuition in \Cref{sec:methods}, to adapt the objective for the ADIOS framework, we apply the masks learned by the occlusion model to one of the views.
We can therefore arrive at our final objective,
\begin{align}
    \mathcal{L}_{\text{SimSiam}}^{\text{ADIOS}}(\x; \I,\M) =  \frac{1}{2}\left(\mathcal{D}(\z^{A,m}, \h^B) + \mathcal{D}(\z^{B,m}, \h^A) \right), \label{eq:simsiam_detailed}
\end{align}
where $\z^{*,m}=g(f(\x^{*,m}))$.

\subsection{BYOL}
\begin{wrapfigure}[13]{r}{0.55\linewidth}
  \vspace{-1.5\baselineskip}
  \includegraphics[width=\linewidth]{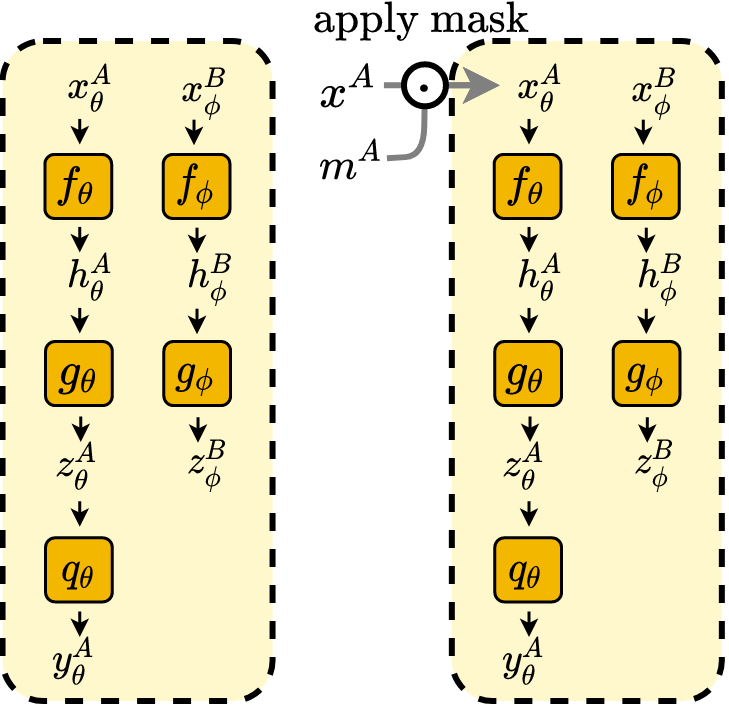}
  \vspace{-1.5\baselineskip}
  \caption{\textbf{\emph{Left}}:~BYOL. \textbf{\emph{Right}}: BYOL + ADIOS.}
  \label{fig:byol_detailed}
\end{wrapfigure}
BYOL \citep{grill2020byol} is another SSL method that avoids the need of negative examples by performing an iterative online update.
Similar to SimSiam, BYOL also adopts an assymetrical forward pass, however different from other approahces, the networks for two different augmentations do not share weights.
See \Cref{fig:byol_detailed} for a visualisation.

For the sake of clarity, we denote the parametrisation of the two networks as $\theta$ and $\phi$.
The $\theta$ network is appended with an additional ``predictor'' $q_\theta$, and is updated via gradient descent using the following objective, which is evaluated using the output of the $\theta$ network $y_\theta$ and output of the $\phi$ network $z_\phi$
\begin{align}
    \mathcal{L}_{\text{BYOL}}(\x; \theta) = \frac{1}{2}\left(\mathcal{D}(\y^A_\theta, \z^B_\phi) + \mathcal{D}(\y^B_\theta, \z^A_\phi) \right),
\end{align}
where $\mathcal{D}$ denotes the mean squared error.
Again, the objective is the average between the two terms due to the assymetrical architecture.

On the other hand, $\phi$ is optimised using the following update rule
\begin{align}
    \phi \leftarrow \tau \phi + (1-\tau) \phi, \label{eq:phi}
\end{align}
where $\tau\in [0,1)$ controls the smoothness of the update. 

To develop the ADIOS objective for BYOL, let us denote $\I$ as the composition of the two networks $\{\I_\theta, \I_\phi\}$.
We can then write
\begin{align}
    \mathcal{L}_{\text{BYOL}}^{\text{ADIOS}}(\x; \I_\theta, \M) = \frac{1}{2}\left(\mathcal{D}(\y^{A,m}_\theta, \z^B_\phi) + \mathcal{D}(\y^{B,m}_\theta, \z^A_\phi) \right),
\end{align}
where $\y_\theta^{*,m} = q_\theta(g_\theta(f_\theta(x^{*,m})))$.
Both $\I_\theta$ and $\M$ are optimised through the min-max objective in \Cref{eq:minmax_obj_n}, whereas $\I_\phi$ is updated by \Cref{eq:phi}.

\section{Backbone of Occlusion Model}\label{sec:app_mask_arch}
We use U-Net \citep{unet} as the backbone of the occlusion model, which is commonly used for semantic segmentation. 
The model consists of a downsampling network, an MLP, and an upsampling network.
We further apply an occlusion head layer with 1x1 kernel to map the output of U-Net to $N$ masks.
Refer to \Cref{tab:unet} for the architecture we used for our experiments.

\begin{table}[t]
\centering
\scalebox{0.8}{%
\begin{tabular}{l}
    \toprule
    \textbf{Down}                                       \\
    \midrule
    3x3 conv.  8 stride 1 pad 1 \& GroupNorm \& ReLU\\
    3x3 conv.  8 stride 1 pad 1 \& GroupNorm \& ReLU\\
    3x3 conv.  16 stride 1 pad 1 \& GroupNorm \& ReLU\\
    3x3 conv.  16 stride 1 pad 1 \& GroupNorm \& ReLU\\
    3x3 conv.  16 stride 1 pad 1 \& GroupNorm \& ReLU\\
    \bottomrule
    \toprule
    \textbf{MLP}                                      \\
    \midrule
    F.C. 128 \& ReLU\\
    F.C. 128 \& ReLU\\
    F.C. 256 \& ReLU\\
    \bottomrule
    \toprule
    \textbf{Up}                                      \\
    \midrule
    3x3 conv.  16 stride 1 pad 1 \& GroupNorm \& ReLU\\
    3x3 conv.  16 stride 1 pad 1 \& GroupNorm \& ReLU\\
    3x3 conv.  8 stride 1 pad 1 \& GroupNorm \& ReLU\\
    3x3 conv.  8 stride 1 pad 1 \& GroupNorm \& ReLU\\
    3x3 conv.  8 stride 1 pad 1 \& GroupNorm \& ReLU\\
    \bottomrule
    \toprule
    \textbf{Occlusion Head}                          \\
    \midrule
    1x1 conv.  $N$ stride 1 pad 1 \& SoftMax\\
    \bottomrule
\end{tabular}}
\caption{U-Net Architecture.}
\label{tab:unet}
\end{table}

\section{Setups of Classificiation Tasks}\label{sec:app_hyper}
We develop our model using solo-learn \citep{sololearn}, a library for state-of-the-art self-supervised learning methods.
For backbones we use ResNet-18 \citep{resnet} and \vit-Tiny \citep{dosovitskiy2021vit} with patch size 16x16.
Hyperparameters including optimiser, momentum, scheduler, epochs and batch size are shared across all models, as seen in \Cref{tab:shared_hp}.
We perform hyperparameter search on the learning rate of encoder for all models, and we include the optimal values used to generated the reported results in \Cref{tab:lr_base};
for the ADIOS models we also run search for the learning rate of the occlusion model, the penalty scaling $\lambda$ and number of masks $N$.
Refer to \Cref{tab:adios_hp} for the values used for these parameters.

\begin{table}[t]
\caption{Hyperparameters for pretraining, used for all models.}\label{tab:shared_hp} \vspace{5pt}
\centering
\scalebox{0.8}{
\begin{tabular}{lc}    \toprule
    Name & Value \\ \midrule
    Optimiser & SGD\\
    Momentum & 0.9\\
    Scheduler & warmup cosine \\
    Epochs & 500 \\
    Batch size & 128\\
    \bottomrule
\end{tabular}}
\end{table}

\begin{table}[t]
\caption{Learning rates for SimCLR, SimSiam and BYOL.}\label{tab:lr_base} \vspace{5pt}
\centering
\scalebox{0.7}{
\begin{tabular}{lcccc}    \toprule
    Architecture & Dataset      & SimCLR & SimSiam & BYOL\\ \midrule
    ResNet18     & ImageNet100-S& 0.15   &  0.25   & 0.25  \\
    ResNet18     & STL10        & 0.15   &  0.23   & 0.31      \\
    \vit-Tiny    & ImageNet100-S& 0.15   &  0.25   & 0.25      \\
    \vit-Tiny    & STL10        & 0.15   &  0.11   & 0.23      \\
    \bottomrule
\end{tabular}}
\end{table}

\begin{table}[t]
\caption{ADIOS hyperparameters for classification tasks.}\label{tab:adios_hp} \vspace{5pt}
\centering
\scalebox{0.75}{
\begin{tabular}{lccc}    \toprule
    & SimCLR+ADIOS & SimSiam+ADIOS & BYOL+ADIOS\\ \midrule
    \multicolumn{3}{l}{\textit{Dataset: ImageNet100-S, backbone: ResNet18}}\\\vspace{2pt}
    Enc. lr & 0.13 & 0.85 & 0.24  \\
    Occ. lr &  0.02  & 0.08 & 0.07 \\
    $\lambda$   &  0.57  & 0.29 & 0.40 \\
    N & 4 & 4 & 4 \\ \midrule
    \multicolumn{3}{l}{\textit{Dataset: ImageNet100-S, backbone: \vit-Tiny}}\\\vspace{2pt}
    Enc. lr & 0.12 & 0.50 & 0.21  \\
    Occ. lr & 0.03 & 0.07 & 0.33  \\
    $\lambda$   & 0.89 & 0.72 & 0.95  \\
    N & 4 & 4 & 4 \\ \midrule
    \multicolumn{3}{l}{\textit{Dataset: STL10, backbone: ResNet18}}\\\vspace{2pt}
    Enc. lr & 0.21 & 0.52 & 0.49  \\
    Occ. lr & 0.33 & 0.29 & 0.06  \\
    $\lambda$   & 0.29 & 0.79 & 0.72  \\
    N & 6 & 6 & 6 \\ \midrule
    \multicolumn{3}{l}{\textit{Dataset: STL10, backbone: \vit-Tiny}}\\\vspace{2pt}
    Enc. lr & 0.14 & 0.56 & 0.29  \\
    Occ. lr & 0.09 & 0.09 & 0.60  \\
    $\lambda$   & 0.50 & 0.12 & 0.18  \\
    N & 6 & 6 & 6 \\ 
    \bottomrule
\end{tabular}}
\end{table}

\section{Setups of Transfer Learning}\label{sec:app_hyper_transfer}
We fine-tune all the models using SGD with a momentum of 0.9 with cosine learning rate decay.
Following protocol in \citet{dosovitskiy2021vit}, we use batch size 512 and no weight decay.
We also run a small grid search on the learning rate with values including $\{0.001, 0.003, 0.01, 0.03\}$.

\section{Setups of CLEVR}\label{sec:app_hyper_clevr}
CLEVR \citep{clevr} is a dataset of rendered 3D objects.
The dataset contains detailed attributes for each object, including shape, color, position, rotation, texture as well as mask, and is commonly used in visual question answering and multi-object representation learning.
Utilising the rich annotations of CLEVR, we construct a challenging multi-label classification task, which we use to evaluate the quality of representations learned under different masking scheme.

For hyperparameters including optimiser, momentum, scheduler, epochs and batch size, we follow the same setup in \Cref{tab:shared_hp}.
We perform hyperparameter search on the learning rate of encoder and occlusion model, as well as the penalty scaling $\lambda$ and number of masks $N$, which we list in \Cref{tab:clevr_hp}.

\begin{table}[t]
\caption{Hyperparameters used for CLEVR.}\label{tab:clevr_hp} \vspace{5pt}
\centering
\scalebox{0.7}{
\begin{tabular}{lcccccc}    \toprule
    & SimCLR & +ADIOS & SimSiam & +ADIOS & BYOL & +ADIOS\\ 
    \cmidrule(lr){2-3} \cmidrule(lr){4-5} \cmidrule(lr){6-7}
    Enc. lr & 0.2 & 0.3 & 0.7  & 0.5 & 0.5  & 0.4 \\
    Occ. lr & - & 0.1 & - &  0.1 & -  & 0.3\\
    $\lambda$   &  - & 0.2  & - & 0.8 & -  & 0.9\\
    N & - & 4 &- & 4 &- & 4 \\ 
    \bottomrule
\end{tabular}}
\end{table}

\section{CIFAR ResNet} \label{sec:cifar_resnet}
As we mentioned, the authors of ResNet \citep{resnet} proposes a slightly different architecture for CIFAR due to their small image size.
The difference between the CIFAR-ResNet and standard ResNet lies in the first convolutional block (before \texttt{layer1}), which we provide a side by side comparison of in \Cref{tab:standard_resnet} and \Cref{tab:cifar_resnet}.
\begin{table}[!htb]
    \begin{minipage}{0.48\linewidth}
    \caption{Standard ResNet.}\label{tab:standard_resnet}
    \centering
    \scalebox{0.8}{%
    \begin{tabular}{l}
        \toprule
        7x7 conv.  64 stride 2 pad 3 \\ 
        BatchNorm\\
        ReLU\\
        3x3 MaxPool stride 2 pad 1\\
        \bottomrule
    \end{tabular}}
    \end{minipage}
    \begin{minipage}{0.48\linewidth}
    \caption{CIFAR ResNet.}\label{tab:cifar_resnet}
    \centering
    \scalebox{0.8}{%
    \begin{tabular}{l}
        \toprule
        3x3 conv.  64 stride 1 pad 2\\ 
        BatchNorm \\
        ReLU\\
        - \\
        \bottomrule
    \end{tabular}}
    \end{minipage}
\end{table}

As we see, the kernel size of the convolutional layer is smaller and the MaxPool operation is removed to suit the small image size.
We adopt this CIFAR-optimal ResNet for fine-tuning, but stick to the original architecture for linear evaluation to utilise the weights learned during pre-training, which resulted in the considerble underperformance on this metric for all 6 models.




%% file: main.bbl
\begin{thebibliography}{39}
\providecommand{\natexlab}[1]{#1}
\providecommand{\url}[1]{\texttt{#1}}
\expandafter\ifx\csname urlstyle\endcsname\relax
  \providecommand{\doi}[1]{doi: #1}\else
  \providecommand{\doi}{doi: \begingroup \urlstyle{rm}\Url}\fi

\bibitem[Baevski et~al.()Baevski, Hsu, Xu, Babu, Gu, and Auli]{data2vec}
Baevski, A., Hsu, W.-N., Xu, Q., Babu, A., Gu, J., and Auli, M.
\newblock Data2vec: A general framework for self-supervised learning in speech,
  vision and language.
\newblock URL
  \url{https://ai.facebook.com/research/data2vec-a-general-framework-for-self
  -supervised-learning-in-speech-vision-and-language/}.
\newblock Accessed: 2022-01-27.

\bibitem[Bao et~al.(2021)Bao, Dong, and Wei]{bao2021beit}
Bao, H., Dong, L., and Wei, F.
\newblock Beit: Bert pre-training of image transformers.
\newblock \emph{ArXiv preprint}, abs/2106.08254, 2021.
\newblock URL \url{https://arxiv.org/abs/2106.08254}.

\bibitem[{Bardes} et~al.(2021){Bardes}, {Ponce}, and {LeCun}]{bardes2021vicreg}
{Bardes}, A., {Ponce}, J., and {LeCun}, Y.
\newblock Vicreg: Variance-invariance-covariance regularization for
  self-supervised learning.
\newblock \emph{ArXiv preprint}, abs/2105.04906, 2021.
\newblock URL \url{https://arxiv.org/abs/2105.04906}.

\bibitem[{Bromley} et~al.(1993){Bromley}, {Bentz}, {Bottou}, {Guyon}, {Lecun},
  {Moore}, {Säckinger}, and {Shah}]{bromley1993siamese}
{Bromley}, J., {Bentz}, J.~W., {Bottou}, L., {Guyon}, I., {Lecun}, Y., {Moore},
  C., {Säckinger}, E., and {Shah}, R.
\newblock Signature verification using a “siamese” time delay neural
  network.
\newblock \emph{International Journal of Pattern Recognition and Artificial
  Intelligence}, 7\penalty0 (4):\penalty0 669--688, 1993.

\bibitem[Caron et~al.(2020)Caron, Misra, Mairal, Goyal, Bojanowski, and
  Joulin]{caron2020swav}
Caron, M., Misra, I., Mairal, J., Goyal, P., Bojanowski, P., and Joulin, A.
\newblock Unsupervised learning of visual features by contrasting cluster
  assignments.
\newblock In Larochelle, H., Ranzato, M., Hadsell, R., Balcan, M., and Lin, H.
  (eds.), \emph{Advances in Neural Information Processing Systems 33: Annual
  Conference on Neural Information Processing Systems 2020, NeurIPS 2020,
  December 6-12, 2020, virtual}, 2020.
\newblock URL
  \url{https://proceedings.neurips.cc/paper/2020/hash/70feb62b69f16e0238f741fab228fec2-Abstract.html}.

\bibitem[Chen et~al.(2020)Chen, Kornblith, Norouzi, and Hinton]{chen2020simclr}
Chen, T., Kornblith, S., Norouzi, M., and Hinton, G.~E.
\newblock A simple framework for contrastive learning of visual
  representations.
\newblock In \emph{Proceedings of the 37th International Conference on Machine
  Learning, {ICML} 2020, 13-18 July 2020, Virtual Event}, volume 119 of
  \emph{Proceedings of Machine Learning Research}, pp.\  1597--1607. {PMLR},
  2020.
\newblock URL \url{http://proceedings.mlr.press/v119/chen20j.html}.

\bibitem[{Chen} \& {He}(2021){Chen} and {He}]{chen2021simsiam}
{Chen}, X. and {He}, K.
\newblock Exploring simple siamese representation learning.
\newblock In \emph{Proceedings of the IEEE/CVF Conference on Computer Vision
  and Pattern Recognition}, pp.\  15750--15758, 2021.

\bibitem[{Chen} et~al.(2020){Chen}, {Fan}, {Girshick}, and
  {He}]{chen2020mocov2}
{Chen}, X., {Fan}, H., {Girshick}, R.~B., and {He}, K.
\newblock Improved baselines with momentum contrastive learning.
\newblock \emph{ArXiv preprint}, abs/2003.04297, 2020.
\newblock URL \url{https://arxiv.org/abs/2003.04297}.

\bibitem[Cubuk et~al.(2019)Cubuk, Zoph, Man{\'{e}}, Vasudevan, and
  Le]{cubuk2019autoaugment}
Cubuk, E.~D., Zoph, B., Man{\'{e}}, D., Vasudevan, V., and Le, Q.~V.
\newblock Autoaugment: Learning augmentation strategies from data.
\newblock In \emph{{IEEE} Conference on Computer Vision and Pattern
  Recognition, {CVPR} 2019, Long Beach, CA, USA, June 16-20, 2019}, pp.\
  113--123. Computer Vision Foundation / {IEEE}, 2019.
\newblock \doi{10.1109/CVPR.2019.00020}.
\newblock URL
  \url{http://openaccess.thecvf.com/content\_CVPR\_2019/html/Cubuk\_AutoAugment\_Learning\_Augmentation\_Strategies\_From\_Data\_CVPR\_2019\_paper.html}.

\bibitem[da~Costa et~al.(2021)da~Costa, Fini, Nabi, Sebe, and Ricci]{sololearn}
da~Costa, V. G.~T., Fini, E., Nabi, M., Sebe, N., and Ricci, E.
\newblock Solo-learn: A library of self-supervised methods for visual
  representation learning, 2021.
\newblock URL \url{https://github.com/vturrisi/solo-learn}.

\bibitem[Devlin et~al.(2019)Devlin, Chang, Lee, and Toutanova]{devlin2018bert}
Devlin, J., Chang, M.-W., Lee, K., and Toutanova, K.
\newblock {BERT}: Pre-training of deep bidirectional transformers for language
  understanding.
\newblock In \emph{Proceedings of the 2019 Conference of the North {A}merican
  Chapter of the Association for Computational Linguistics: Human Language
  Technologies, Volume 1 (Long and Short Papers)}, pp.\  4171--4186,
  Minneapolis, Minnesota, 2019. Association for Computational Linguistics.
\newblock \doi{10.18653/v1/N19-1423}.
\newblock URL \url{https://aclanthology.org/N19-1423}.

\bibitem[Dosovitskiy et~al.(2021)Dosovitskiy, Beyer, Kolesnikov, Weissenborn,
  Zhai, Unterthiner, Dehghani, Minderer, Heigold, Gelly, Uszkoreit, and
  Houlsby]{dosovitskiy2021vit}
Dosovitskiy, A., Beyer, L., Kolesnikov, A., Weissenborn, D., Zhai, X.,
  Unterthiner, T., Dehghani, M., Minderer, M., Heigold, G., Gelly, S.,
  Uszkoreit, J., and Houlsby, N.
\newblock An image is worth 16x16 words: Transformers for image recognition at
  scale.
\newblock In \emph{9th International Conference on Learning Representations,
  {ICLR} 2021, Virtual Event, Austria, May 3-7, 2021}. OpenReview.net, 2021.
\newblock URL \url{https://openreview.net/forum?id=YicbFdNTTy}.

\bibitem[Engelcke et~al.(2020)Engelcke, Kosiorek, Jones, and
  Posner]{engelcke2019genesis}
Engelcke, M., Kosiorek, A.~R., Jones, O.~P., and Posner, I.
\newblock {GENESIS:} generative scene inference and sampling with
  object-centric latent representations.
\newblock In \emph{8th International Conference on Learning Representations,
  {ICLR} 2020, Addis Ababa, Ethiopia, April 26-30, 2020}. OpenReview.net, 2020.
\newblock URL \url{https://openreview.net/forum?id=BkxfaTVFwH}.

\bibitem[{Engelcke} et~al.(2021){Engelcke}, {Jones}, and
  {Posner}]{engelcke2021genesis}
{Engelcke}, M., {Jones}, O.~P., and {Posner}, I.
\newblock Genesis-v2: Inferring unordered object representations without
  iterative refinement.
\newblock \emph{ArXiv preprint}, abs/2104.09958, 2021.
\newblock URL \url{https://arxiv.org/abs/2104.09958}.

\bibitem[Ermolov et~al.(2021)Ermolov, Siarohin, Sangineto, and
  Sebe]{ermolov2021wmse}
Ermolov, A., Siarohin, A., Sangineto, E., and Sebe, N.
\newblock Whitening for self-supervised representation learning.
\newblock In Meila, M. and Zhang, T. (eds.), \emph{Proceedings of the 38th
  International Conference on Machine Learning, {ICML} 2021, 18-24 July 2021,
  Virtual Event}, volume 139 of \emph{Proceedings of Machine Learning
  Research}, pp.\  3015--3024. {PMLR}, 2021.
\newblock URL \url{http://proceedings.mlr.press/v139/ermolov21a.html}.

\bibitem[{Gansbeke} et~al.(2020){Gansbeke}, {Vandenhende}, {Georgoulis},
  {Proesmans}, and {Gool}]{gansbeke2020scan}
{Gansbeke}, W.~V., {Vandenhende}, S., {Georgoulis}, S., {Proesmans}, M., and
  {Gool}, L.~V.
\newblock Scan: Learning to classify images without labels.
\newblock In \emph{ECCV (10)}, pp.\  268--285, 2020.

\bibitem[Greff et~al.(2019)Greff, Kaufman, Kabra, Watters, Burgess, Zoran,
  Matthey, Botvinick, and Lerchner]{greff2019iodine}
Greff, K., Kaufman, R.~L., Kabra, R., Watters, N., Burgess, C., Zoran, D.,
  Matthey, L., Botvinick, M., and Lerchner, A.
\newblock Multi-object representation learning with iterative variational
  inference.
\newblock In Chaudhuri, K. and Salakhutdinov, R. (eds.), \emph{Proceedings of
  the 36th International Conference on Machine Learning, {ICML} 2019, 9-15 June
  2019, Long Beach, California, {USA}}, volume~97 of \emph{Proceedings of
  Machine Learning Research}, pp.\  2424--2433. {PMLR}, 2019.
\newblock URL \url{http://proceedings.mlr.press/v97/greff19a.html}.

\bibitem[Grill et~al.(2020)Grill, Strub, Altch{\'{e}}, Tallec, Richemond,
  Buchatskaya, Doersch, Pires, Guo, Azar, Piot, Kavukcuoglu, Munos, and
  Valko]{grill2020byol}
Grill, J., Strub, F., Altch{\'{e}}, F., Tallec, C., Richemond, P.~H.,
  Buchatskaya, E., Doersch, C., Pires, B.~{\'{A}}., Guo, Z., Azar, M.~G., Piot,
  B., Kavukcuoglu, K., Munos, R., and Valko, M.
\newblock Bootstrap your own latent - {A} new approach to self-supervised
  learning.
\newblock In Larochelle, H., Ranzato, M., Hadsell, R., Balcan, M., and Lin, H.
  (eds.), \emph{Advances in Neural Information Processing Systems 33: Annual
  Conference on Neural Information Processing Systems 2020, NeurIPS 2020,
  December 6-12, 2020, virtual}, 2020.
\newblock URL
  \url{https://proceedings.neurips.cc/paper/2020/hash/f3ada80d5c4ee70142b17b8192b2958e-Abstract.html}.

\bibitem[{Hataya} et~al.(2020){Hataya}, {Zdenek}, {Yoshizoe}, and
  {Nakayama}]{hataya2020fasterautoaugment}
{Hataya}, R., {Zdenek}, J., {Yoshizoe}, K., and {Nakayama}, H.
\newblock Faster autoaugment: Learning augmentation strategies using
  backpropagation.
\newblock In \emph{16th European Conference on Computer Vision, ECCV 2020},
  pp.\  1--16, 2020.

\bibitem[He et~al.(2016)He, Zhang, Ren, and Sun]{resnet}
He, K., Zhang, X., Ren, S., and Sun, J.
\newblock Deep residual learning for image recognition.
\newblock In \emph{2016 {IEEE} Conference on Computer Vision and Pattern
  Recognition, {CVPR} 2016, Las Vegas, NV, USA, June 27-30, 2016}, pp.\
  770--778. {IEEE} Computer Society, 2016.
\newblock \doi{10.1109/CVPR.2016.90}.
\newblock URL \url{https://doi.org/10.1109/CVPR.2016.90}.

\bibitem[He et~al.(2020)He, Fan, Wu, Xie, and Girshick]{he2020moco}
He, K., Fan, H., Wu, Y., Xie, S., and Girshick, R.~B.
\newblock Momentum contrast for unsupervised visual representation learning.
\newblock In \emph{2020 {IEEE/CVF} Conference on Computer Vision and Pattern
  Recognition, {CVPR} 2020, Seattle, WA, USA, June 13-19, 2020}, pp.\
  9726--9735. {IEEE}, 2020.
\newblock \doi{10.1109/CVPR42600.2020.00975}.
\newblock URL \url{https://doi.org/10.1109/CVPR42600.2020.00975}.

\bibitem[He et~al.(2021)He, Chen, Xie, Li, Dollár, and Girshick]{he2021mae}
He, K., Chen, X., Xie, S., Li, Y., Dollár, P., and Girshick, R.
\newblock Masked autoencoders are scalable vision learners.
\newblock \emph{ArXiv}, abs/2111.06377, 2021.

\bibitem[Horn et~al.(2018)Horn, Aodha, Song, Cui, Sun, Shepard, Adam, Perona,
  and Belongie]{inaturalist}
Horn, G.~V., Aodha, O.~M., Song, Y., Cui, Y., Sun, C., Shepard, A., Adam, H.,
  Perona, P., and Belongie, S.~J.
\newblock The inaturalist species classification and detection dataset.
\newblock In \emph{2018 {IEEE} Conference on Computer Vision and Pattern
  Recognition, {CVPR} 2018, Salt Lake City, UT, USA, June 18-22, 2018}, pp.\
  8769--8778. {IEEE} Computer Society, 2018.
\newblock \doi{10.1109/CVPR.2018.00914}.
\newblock URL
  \url{http://openaccess.thecvf.com/content\_cvpr\_2018/html/Van\_Horn\_The\_INaturalist\_Species\_CVPR\_2018\_paper.html}.

\bibitem[Johnson et~al.(2017)Johnson, Hariharan, van~der Maaten, Fei{-}Fei,
  Zitnick, and Girshick]{clevr}
Johnson, J., Hariharan, B., van~der Maaten, L., Fei{-}Fei, L., Zitnick, C.~L.,
  and Girshick, R.~B.
\newblock {CLEVR:} {A} diagnostic dataset for compositional language and
  elementary visual reasoning.
\newblock In \emph{2017 {IEEE} Conference on Computer Vision and Pattern
  Recognition, {CVPR} 2017, Honolulu, HI, USA, July 21-26, 2017}, pp.\
  1988--1997. {IEEE} Computer Society, 2017.
\newblock \doi{10.1109/CVPR.2017.215}.
\newblock URL \url{https://doi.org/10.1109/CVPR.2017.215}.

\bibitem[Koyama et~al.(2021)Koyama, Minami, Miyato, and
  Gal]{koyama2021contrastive}
Koyama, M., Minami, K., Miyato, T., and Gal, Y.
\newblock Contrastive representation learning with trainable augmentation
  channel.
\newblock \emph{ArXiv preprint}, abs/2111.07679, 2021.
\newblock URL \url{https://arxiv.org/abs/2111.07679}.

\bibitem[Kr{\"{a}}henb{\"{u}}hl \& Koltun(2011)Kr{\"{a}}henb{\"{u}}hl and
  Koltun]{crf}
Kr{\"{a}}henb{\"{u}}hl, P. and Koltun, V.
\newblock Efficient inference in fully connected crfs with gaussian edge
  potentials.
\newblock In Shawe{-}Taylor, J., Zemel, R.~S., Bartlett, P.~L., Pereira, F.
  C.~N., and Weinberger, K.~Q. (eds.), \emph{Advances in Neural Information
  Processing Systems 24: 25th Annual Conference on Neural Information
  Processing Systems 2011. Proceedings of a meeting held 12-14 December 2011,
  Granada, Spain}, pp.\  109--117, 2011.
\newblock URL
  \url{https://proceedings.neurips.cc/paper/2011/hash/beda24c1e1b46055dff2c39c98fd6fc1-Abstract.html}.

\bibitem[Krizhevsky et~al.(2009)Krizhevsky, Hinton, et~al.]{cifar}
Krizhevsky, A., Hinton, G., et~al.
\newblock Learning multiple layers of features from tiny images.
\newblock 2009.

\bibitem[Liu et~al.(2022)Liu, Mao, Chao-Yuan, Feichtenhofer, Darrell, and
  Xie]{2022convnet}
Liu, Z., Mao, H., Chao-Yuan, W., Feichtenhofer, C., Darrell, T., and Xie, S.
\newblock A convnet for the 2020s.
\newblock \emph{arXiv preprint arXiv: 2201.03545}, 2022.

\bibitem[Nilsback \& Zisserman(2008)Nilsback and Zisserman]{flowers}
Nilsback, M.-E. and Zisserman, A.
\newblock Automated flower classification over a large number of classes.
\newblock In \emph{2008 Sixth Indian Conference on Computer Vision, Graphics \&
  Image Processing}, pp.\  722--729. IEEE, 2008.

\bibitem[Pathak et~al.(2016)Pathak, Kr{\"{a}}henb{\"{u}}hl, Donahue, Darrell,
  and Efros]{Pathak2016context}
Pathak, D., Kr{\"{a}}henb{\"{u}}hl, P., Donahue, J., Darrell, T., and Efros,
  A.~A.
\newblock Context encoders: Feature learning by inpainting.
\newblock In \emph{2016 {IEEE} Conference on Computer Vision and Pattern
  Recognition, {CVPR} 2016, Las Vegas, NV, USA, June 27-30, 2016}, pp.\
  2536--2544. {IEEE} Computer Society, 2016.
\newblock \doi{10.1109/CVPR.2016.278}.
\newblock URL \url{https://doi.org/10.1109/CVPR.2016.278}.

\bibitem[Ramesh et~al.(2021)Ramesh, Pavlov, Goh, Gray, Voss, Radford, Chen, and
  Sutskever]{ramesh2021zero}
Ramesh, A., Pavlov, M., Goh, G., Gray, S., Voss, C., Radford, A., Chen, M., and
  Sutskever, I.
\newblock Zero-shot text-to-image generation.
\newblock In Meila, M. and Zhang, T. (eds.), \emph{Proceedings of the 38th
  International Conference on Machine Learning, {ICML} 2021, 18-24 July 2021,
  Virtual Event}, volume 139 of \emph{Proceedings of Machine Learning
  Research}, pp.\  8821--8831. {PMLR}, 2021.
\newblock URL \url{http://proceedings.mlr.press/v139/ramesh21a.html}.

\bibitem[Ronneberger et~al.(2015)Ronneberger, Fischer, and Brox]{unet}
Ronneberger, O., Fischer, P., and Brox, T.
\newblock U-net: Convolutional networks for biomedical image segmentation.
\newblock In \emph{International Conference on Medical image computing and
  computer-assisted intervention}, pp.\  234--241. Springer, 2015.

\bibitem[{Russakovsky} et~al.(2015){Russakovsky}, {Deng}, {Su}, {Krause},
  {Satheesh}, {Ma}, {Huang}, {Karpathy}, {Khosla}, {Bernstein}, {Berg}, and
  {Fei-Fei}]{imagenet}
{Russakovsky}, O., {Deng}, J., {Su}, H., {Krause}, J., {Satheesh}, S., {Ma},
  S., {Huang}, Z., {Karpathy}, A., {Khosla}, A., {Bernstein}, M., {Berg},
  A.~C., and {Fei-Fei}, L.
\newblock Imagenet large scale visual recognition challenge.
\newblock \emph{International Journal of Computer Vision}, 115\penalty0
  (3):\penalty0 211--252, 2015.

\bibitem[Tamkin et~al.(2021)Tamkin, Wu, and Goodman]{tamkin2021viewmaker}
Tamkin, A., Wu, M., and Goodman, N.~D.
\newblock Viewmaker networks: Learning views for unsupervised representation
  learning.
\newblock In \emph{9th International Conference on Learning Representations,
  {ICLR} 2021, Virtual Event, Austria, May 3-7, 2021}. OpenReview.net, 2021.
\newblock URL \url{https://openreview.net/forum?id=enoVQWLsfyL}.

\bibitem[Tian et~al.(2020)Tian, Krishnan, and Isola]{tian2020imagenet100}
Tian, Y., Krishnan, D., and Isola, P.
\newblock Contrastive multiview coding.
\newblock In \emph{Computer Vision--ECCV 2020: 16th European Conference,
  Glasgow, UK, August 23--28, 2020, Proceedings, Part XI 16}, pp.\  776--794.
  Springer, 2020.

\bibitem[Touvron et~al.(2021)Touvron, Cord, El-Nouby, Bojanowski, Joulin,
  Synnaeve, and J{\'e}gou]{touvron2021augmenting}
Touvron, H., Cord, M., El-Nouby, A., Bojanowski, P., Joulin, A., Synnaeve, G.,
  and J{\'e}gou, H.
\newblock Augmenting convolutional networks with attention-based aggregation.
\newblock \emph{ArXiv preprint}, abs/2112.13692, 2021.
\newblock URL \url{https://arxiv.org/abs/2112.13692}.

\bibitem[Xiao et~al.(2021)Xiao, Engstrom, Ilyas, and Madry]{bgchallenge}
Xiao, K.~Y., Engstrom, L., Ilyas, A., and Madry, A.
\newblock Noise or signal: The role of image backgrounds in object recognition.
\newblock In \emph{9th International Conference on Learning Representations,
  {ICLR} 2021, Virtual Event, Austria, May 3-7, 2021}. OpenReview.net, 2021.
\newblock URL \url{https://openreview.net/forum?id=gl3D-xY7wLq}.

\bibitem[Zbontar et~al.(2021)Zbontar, Jing, Misra, LeCun, and
  Deny]{zbontar2021barlow}
Zbontar, J., Jing, L., Misra, I., LeCun, Y., and Deny, S.
\newblock Barlow twins: Self-supervised learning via redundancy reduction.
\newblock In Meila, M. and Zhang, T. (eds.), \emph{Proceedings of the 38th
  International Conference on Machine Learning, {ICML} 2021, 18-24 July 2021,
  Virtual Event}, volume 139 of \emph{Proceedings of Machine Learning
  Research}, pp.\  12310--12320. {PMLR}, 2021.
\newblock URL \url{http://proceedings.mlr.press/v139/zbontar21a.html}.

\bibitem[Zhou et~al.(2021)Zhou, Wei, Wang, Shen, Xie, Yuille, and
  Kong]{zhou2021ibot}
Zhou, J., Wei, C., Wang, H., Shen, W., Xie, C., Yuille, A., and Kong, T.
\newblock ibot: Image bert pre-training with online tokenizer.
\newblock \emph{ArXiv preprint}, abs/2111.07832, 2021.
\newblock URL \url{https://arxiv.org/abs/2111.07832}.

\end{thebibliography}
